\documentclass[11pt]{article}
\usepackage{graphicx}
\usepackage{amssymb}
\usepackage{amscd}
\usepackage{mathrsfs}
\usepackage{longtable,lscape}
\usepackage{amsthm}
\usepackage{amsfonts}
\usepackage{amsmath}
\usepackage{bbm}
\usepackage{float}
\usepackage{subfig}
\usepackage{url}
\usepackage{hyperref}
\usepackage{lineno}
\usepackage{csquotes}
\usepackage{wrapfig}
\usepackage[round]{natbib}
\usepackage{caption}
\usepackage[title]{appendix}
\begin{document}
\title{Identifying Quantum Structure in AI Language: Evidence for Evolutionary Convergence of Human and Artificial Cognition}

\author{Diederik Aerts$^*$, Jonito Aerts Argu{\"e}lles$^*$, Lester Beltran$^*$, \\ Suzette Geriente$^*$, Massimiliano Sassoli de Bianchi\footnote{Center Leo Apostel for Interdisciplinary Studies, Vrije Universiteit Brussel (VUB), Pleinlaan 2, 1050 Brussels, Belgium; email addresses: diraerts@vub.be, jonitoarguelles@gmail.com, lestercc21@yahoo.com, sgeriente83@yahoo.com, autoricerca@gmail.com.}, \\
        Roberto Leporini\footnote{
        Department of Economics, University of Bergamo, via dei Caniana 2, Bergamo, 24127, Italy; email address: roberto.leporini@unibg.it.}
        $\,$ and Sandro Sozzo\footnote{Department of Humanities and Cultural Heritage (DIUM) and Centre CQSCS, University of Udine, Vicolo Florio 2/b, 33100 Udine, Italy; email address: sandro.sozzo@uniud.it.}              }
\date{}
\maketitle

\begin{abstract} 
\noindent 
We present the results of cognitive tests on conceptual combinations, performed using specific Large Language Models (LLMs) as test subjects. In the first test, performed with ChatGPT and Gemini, we show that Bell's inequalities are significantly violated, which indicates the presence of a `non-classical probability model' with probabilities that do not satisfy Kolmogorov's axioms. In the second test, also performed using ChatGPT and Gemini, we identify the presence of `Bose-Einstein statistics', rather than the intuitively expected `Maxwell-Boltzmann statistics', in the distribution of the words contained in large-size texts. Interestingly, these findings mirror the results previously obtained in both cognitive tests with human participants and information retrieval tests on large corpora. Taken together, they point to the `systematic emergence of non-classical quantum-like structures in conceptual-linguistic domains', regardless of whether the cognitive agent is human or artificial. Although LLMs are classified as neural networks for historical reasons, we believe that a more essential form of knowledge organization takes place in the distributive semantic structure of vector spaces built on top of the neural network. It is this meaning-bearing structure that lends itself to a phenomenon of evolutionary convergence between human cognition and language, slowly established through biological evolution, and LLM cognition and language, emerging much more rapidly as a result of self-learning and training. We analyze various aspects and examples that contain evidence supporting the above hypothesis. We also advance a unifying framework that explains the pervasive quantum organization of meaning that we identify.    
\end{abstract} 

\section{Introduction} 

In this study, we performed and analyzed experiments testing Bell's inequalities and quantum statistics in the cognitive domain, using large language models (LLMs) as test subjects. As we shall see, LLMs violate Bell's inequalities and explicitly exhibit quantum statistics in the texts they produce. It is within the research domain known as quantum cognition \citep{aerts2009a, aerts2009b, bruzagabora2009, aertssozzo2011, busemeyerbruza2012, havenkhrennikov2013, aertssozzo2014, dallachiaraetal2015, pothosetal2015, blutnerbeimgraben2016, moreirawichert2016, gaborakitto2017, surovetal2019, aertsbeltran2020, aertsbeltran2022a, PothosBusemeyer2022, huangetal2025}, where structures such as the quantum probability model and the complex Hilbert space of quantum mechanics are used to describe phenomena of human language and decision-making, that we examine the current capabilities of LLMs. This rapidly expanding framework has recently seen profound developments, moving from purely behavioral models of decision anomalies to comprehensive structural overviews of the entire research program \citep{huangetal2025}, expansions into episodic memory frameworks \citep{BusemeyerOzawa2025}, applications analyzing comparative reasoning patterns in advanced generative architectures like GPT models \citep{Imannezhad2026}, and new foundational inquiries into perception and the qualia hypothesis \citep{Tsuchiya2025}. By applying these highly active, non-classical probabilistic tools to LLMs as test subjects, we aim to highlight surprising aspects of artificial cognitive capabilities that remain completely underexposed in conventional machine learning literature.

In 1964, John Bell proved that, if one introduces local realism as a hypothesis for a physical theory, then one can derive inequalities for the expectation values of suitable physical observables (Bell's inequalities) which are violated in quantum mechanics \cite{bell1964}. This violation is due to a feature of quantum mechanics which is called `entanglement'. However, in our analysis of the violation of Bell’s inequalities by LLM's, we will draw on Itamar Pitowsky’s work on `correlation polytopes' \citep{pitowsky1989}, which in turn is based on George Boole’s original work on `conditions of possible experience' \citep{Boole1854}. Pitowsky demonstrates that a violation of the Clauser-Horne-Shimoni-Holt  (CHSH) version of Bell’s inequalities demonstrates the presence of a non-Kolmogorovian probability model underlying the formation of the identified correlations \citep{kolmogorov1932,pitowsky1989}. In his later work, Pitowsky also demonstrates that the CHSH inequalities can be derived from the mathematical expressions proposed by George Boole in his original analysis of human cognition \citep{Pitowsky1994,Boole1854}.
One generally concludes that, because of entanglement, one cannot consider the component parts of a composite quantum system, or `entity', separately, but the entity must be described as an undivided whole. The way in which Bell's inequalities are violated and how this is proven by the tested LLMs touches on all three of the above cases (entanglement, non-Kolmogorovianity and wholeness) but is also specific to how quantum structures emerge in cognition and the information exchange that accompanies it, as we will clarify in the next section. 

When it comes to large collections of quantum entities, be they molecules, atoms, photons, or other quantum entities, they always behave according to one of two quantum statistics, Bose-Einstein or Fermi-Dirac, and never according to Maxwell-Boltzmann statistics, the statistics that govern collections of classical entities. We were therefore surprised when, five years ago, our investigation of the statistical structure of collections of words in human-written stories revealed that it is clearly Bose-Einstein and not Maxwell-Boltzmann \citep{aertsbeltran2020}. Although we had suspected this to be the case based on previous research \citep{aerts2009b,aertssozzoveloz2015b,beltran2023}, we were still not expecting that the Bose-Einstein statistics described the distribution of words in a story so perfectly, as we demonstrated in \citet{aertsbeltran2020}. What about the collections of words generated by LLMs? Since LLMs are self-learning and use in their training the texts available on the Web, which are largely written by humans, it is to be expected that stories written by LLMs will also obey the Bose-Einstein statistics, as was indeed confirmed in our study. As we will explain in the article, this presence of Bose-Einstein statistics in the output of the LLMs indicates the existence of a non-classical structure underlying their code. This will allow us to draw conclusions about the nature and depth of cognition and the associated information exchange.

Our article is organized as follows. In Section ~\ref{LLMBell}, we demonstrate that the results of a specific set of cognitive tests performed with ChatGPT and Gemini as test subjects violate the CHSH inequalities. We do this in two stages. First, the LLMs act as a single subject and, during a conversation, are guided through the various questions contained in the four tests they are subjected to, step by step. This leads the LLMs to generate the most likely answers, thereby maximally violating the CHSH inequalities. This procedure is described in Appendix~\ref{appendixa}. In the second stage, we proceed in a manner more similar to experiments conducted with human subjects. This time, we ask ChatGPT and Gemini to perform the four tests $81$ times, as we did in experiments with human subjects \citet{aertssozzo2011}, using two different prompts. For the first prompt, we instruct the LLMs to provide exploratory responses; for the second, we simply omit this instruction. This results in violations that are no longer maximal, but closer to the range of values observed in experiments with human subjects. The two prompts are described in Appendix~\ref{appendixb}.

In Section~\ref{BEandstories}, we demonstrate that the statistical structure of the stories written by ChatGPT and Gemini is of Bose-Einstein type and not of Maxwell-Boltzmann type. The way we proceeded is as follows. After explaining the nature of our research to the LLMs, we asked them if they would like to participate in our study by writing a story in the style of A. Milne, using the Winnie the Pooh characters as a basis, without any suggestions regarding the content of the story. Both LLMs responded enthusiastically and produced a Winnie the Pooh story. We did not guide the content, but had to clarify that the story would be their own original output. Note that the AIs initially claimed that they would need a few days but then failed to deliver. This reflects a typical case of hallucination, since AIs have no internal sense of time. In addition, the subsequent failure to provide the promised output is more plausibly explained by constraints such as copyright restrictions, tool malfunctions, or other systemic limits that cause the model to terminate silently. We specifically asked them to write a Winnie the Pooh story because it was precisely a Winnie the Pooh story that we had first analyzed, from a statistical point of view, discovering to our surprise that it obeyed Bose-Einstein statistics and not Maxwell-Boltzmann statistics, as we had originally expected \citep{aertsbeltran2020}. Gemini also wrote a second story in the style of H. G. Wells, which we also report on in Section \ref{BEandstories}, and which, after analysis, also clearly exhibited the Bose-Einstein statistical structure. 

We found the three stories resulting from our experiment to be beautiful and well-written. Apparently, an attempt had been made to give our research on the connections between human- and artificially-generated languages, and quantum structures, an artistic dimension.
The Winnie the Pooh story written by Gemini is reproduced in its entirety in the  
Appendix~\ref{appendixc}. The other two stories are included as supplementary material. In Section~\ref{BEandstories}, we also present hypotheses related to our findings, which reveal the presence of a non-classical cognitive structure in the LLMs code, which they acquire through self-learning and training. We discuss what these results mean for the cognitive nature of LLMs and the emerging phenomenon of AI in general.

In Section~\ref{eunification}, we broaden the perspective of our research by relating it to the unification of human language, AI language, and quanta of light, based on our earlier work. Inspired by Martin Klein’s observation that Einstein sought to unify disparate domains -- Maxwell’s electromagnetism and Boltzmann’s thermodynamics -- through his then counter-current view of light quanta as a gas of atoms, we highlight the parallel importance of our own inquiry. History shows that serious attempts at unification often yield profound breakthroughs and fundamental insights. Following this perspective, we explore the convergence of human and AI-generated language with quanta of light, providing some examples.

\section{LLMs that violate Bell's inequalities indicating the presence of non-classical probability \label{LLMBell}}

In previous work, we studied the violation of Bell’s inequalities in human cognition by developing specific examples of tests administered to a group of participants and observed a clear violation of the inequalities \citep{aertssozzo2011,aertssozzo2014,aertsetal2019,aertsetal2021,aertsetal2025c}. In the present study, we focus on one of these test examples, with LLMs replacing human subjects. 

The violation of Bell's inequalities in physics coincided with one of the most profound upheavals in the history of modern science, as one of the deepest intuitions about the nature of reality was called into question. This inquiry began with the seminal challenge posed by Einstein, Podolsky, and Rosen \citep{EPR1935}, who questioned the completeness of quantum mechanics based on the principle of `local realism'. In 1964, John Bell mathematically formalized this tension by proving that no theory founded on `local hidden variables' could reproduce the correlations predicted by quantum theory. In doing so, he formulated inequalities that test whether a theory satisfies such local hidden variables. These inequalities are violated by quantum mechanics \citep{bell1964}. The subsequent  verification of whether experiments follow the quantum predictions and hence violate Bell's inequalities  was achieved through a series of landmark milestones, beginning with the work of John Clauser \citep{Clauser1972}, followed by the refined experiments of Alain Aspect (\citeyear{Aspect1982}), and extending to the multi-particle entanglement studies of Anton Zeilinger \citep{Zeilinger1998}. The definitive role of these three pioneers in demonstrating the violation of Bell's inequalities was recognized when they were jointly awarded the 2022 Nobel Prize in Physics.

In these physical experiments, the `no-signaling' condition, i.e., the requirement that information cannot travel between measurement sites at speeds exceeding that of light, is the essential criterion for identifying non-locality in spacetime. However, it is essential to distinguish between physical location and semantic context. On a physical level, surprise arises from instantaneous connections perceived across arbitrarily large spatial distances. In our research on LLM cognition, the surprise lies in the structural organization of meaning within a shared, high-dimensional semantic field. Less well known than the EPR paper and the subsequent experimental tests of Bell's inequalities in relation to non-locality are the two papers in which Erwin Schr\"odinger examined a specific aspect of the situation: the fact that, in composite quantum entities, subsystems do not exist as completely independent parts of the whole \citep{schrodinger1935a,schrodinger1935b}. Schr\"odinger also coined the term ``entanglement'' to describe this phenomenon. When we examine more closely what occurs when Bell inequalities are violated by humans and LLMs, it is primarily this Schr\"odingerian phenomenon that is at work. Words, when combined, behave in ways that differ fundamentally from how they behave in isolation. This is why quantum cognition uses the term ``entanglement'' to describe the non-classical manner in which words combine. In \citet{aertssozzo2014}, we further show that the relevant data can be modeled in a tensor product Hilbert space by means of non-product states or non-product self-adjoint operators, that is, in a way that is now mathematically standardly described as a situation of entanglement.

However, we are also interested in identifying, as systematically and precisely as possible, the type of non-classicism at play, which is why we refer to the work of Itamar Pitowsky.
As he demonstrated, a violation of Bell's inequalities provides an explicit proof that the underlying probability model is non-Kolmogorovian \citep{pitowsky1989}. By this, we mean that the observed correlations cannot be contained within the classical axiomatic framework established by Andrey Kolmogorov (\citeyear{kolmogorov1932}), where all events are represented as subsets of a single, global probability space with a well-defined joint distribution.  In a later synthesis, Pitowsky showed that these  were fundamentally anticipated as early as 1854 by George Boole in his investigation into the `laws of thought' \citep{Boole1854,Pitowsky1994}. The `conditions of possible experience' that Boole derived are mathematical boundaries that must be satisfied if a set of experimental data is to be considered logically consistent with classical, rational reasoning. 

In addition to the well-known inequalities reformulated by Bell, researchers in quantum cognition have also tested other conditions for possible experience proposed by Boole to investigate how human decision-making and conceptual combinations deviate from classical logic \citep{bruzaetal2023}. By applying these tests to LLMs, we demonstrate that artificial agents, much like human participants, violated Boole's conditions.

Let us now examine the details of how we investigated violations of Bell’s inequalities in humans and LLMs. Concretely, we asked ChatGPT and Gemini the same questions previously asked to human participants and then analyzed their answers. Before describing the specific tests and the answers provided by ChatGPT and Gemini, we first outline the measurement through which Bell’s inequalities were violated in human cognition, and then explain how we applied an equivalent procedure to LLMs.

We consider the sentence {\it The Animal Acts} as a combination of the concepts {\it Animal} and {\it Acts}. For the concept {\it Animal}, we also consider two pairs of exemplars, or states of the concept {\it Animal}, namely {\it Horse} and {\it Bear}, and {\it Tiger} and {\it Cat}. For the concept {\it Acts}, we also consider two pairs of exemplars, namely {\it Growls} and {\it Whinnies}, and {\it Snorts} and {\it Meows}. Hence, by {\it Acts} we mean here the specific action of {\it Making a Sound}. We then introduce the measurements $A$ and $A'$ for the concept {\it Animal}, and $B$ and $B'$ for the concept {\it Acts}. Measurement $A$ consists of participants choosing between {\it Horse} and {\it Bear}, answering the question `What is a good example of {\it Animal}?' We put as outcome $\lambda_H=+1$ if {\it Horse} is chosen, hence the state of {\it Animal} changes to {\it Horse}, and $\lambda_B=-1$ if {\it Bear} is chosen, hence the state of {\it Animal} changes to {\it Bear}. Measurement 
$A'$ consists of participants choosing between {\it Tiger} and {\it Cat}, answering the question `What is a good example of {\it Animal}?'. We consistently put $\lambda_T=+1$ if {\it Tiger} is chosen and $\lambda_C=-1$ if {\it Cat} is chosen. Measurement 
$B$ consists of participants choosing between {\it Growls} and {\it Whinnies}, answering the question `What is a good example of {\it Acts}?' We put $\lambda_G=+1$ if {\it Growls} is chosen and $\lambda_W=-1$ if {\it Whinnies} is chosen. Measurement $B'$ consists of participants choosing between {\it Snorts} and {\it Meows}, answering the question `What is a good example of {\it Acts}?' Again, we put $\lambda_S=+1$ if {\it Snorts} is chosen and $\lambda_M=-1$ if {\it Meows} is chosen. 

Let us now introduce the coincidence measurements $AB$, $AB'$, $A'B$ and $A'B'$ for the conceptual combination {\it The Animal Acts}. In all measurements, we ask test subjects to answer the question `What is a good example of {\it The Animal Acts}?' In measurement $AB$, participants choose among the following four possibilities, (1) {\it The Horse Growls}, (2) {\it The Bear Whinnies}, (3) {\it The Horse Whinnies}, (4) {\it The Bear Growls}, and we put as outcomes $\lambda_{HG}=+1$, $\lambda_{BW}=+1$, $\lambda_{HW}=-1$, $\lambda_{BG}=-1$, respectively. In measurement $AB'$, they choose among (1) {\it The Horse Snorts}, (2) {\it The Bear Meows}, (3) {\it The Horse Meows}, (4) {\it The Bear Snorts}, and we put as outcomes $\lambda_{HS}=+1$, $\lambda_{BM}=+1$, $\lambda_{HM}=-1$, $\lambda_{BS}=-1$, respectively. In measurement $A'B$, they choose among (1) {\it The Tiger Growls}, (2) {\it The Cat Whinnies}, (3) {\it The Tiger Whinnies}, (4) {\it The Cat Growls}, and we put as outcomes $\lambda_{TG}=+1$, $\lambda_{CW}=+1$, $\lambda_{TW}=-1$, $\lambda_{CG}=-1$, respectively. Finally, in measurement $A'B'$, participants choose among (1) {\it The Tiger Snorts}, (2) {\it The Cat Meows}, (3) {\it The Tiger Meows}, (4)  {\it The Cat Snorts}, and we put as outcomes $\lambda_{TS}=+1$, $\lambda_{CM}=+1$, $\lambda_{TM}=-1$, $\lambda_{CS}=-1$, respectively. 

With the data collected \citep{aertssozzo2011}, we evaluated the expectation values $E(A,B)$, $E(A, B')$, $E(A', B)$ and $E(A', B')$ associated with the measurements $AB$, $AB'$, $A'B$ and $A'B'$, respectively, and inserted the obtained values into the CHSH inequality \citep{clauseretal1969}  
\begin{equation} \label{chsh}
-2 \le E(A',B')+E(A',B)+E(A,B')-E(A,B) \le 2
\end{equation} 
More specifically, the experiment reported in \citet{aertssozzo2011} involved 
$81$ participants who were presented with a questionnaire in which they were asked to make a choice among the above alternatives in the four single-concept measurements $A$, $B$, $A'$ and $B'$, and in the four two-concept coincidence measurements $AB$, $AB'$, $A'B$ and $A'B'$. Table \ref{tab} contains the results. 

If we denote by $P(A_1,B_1)$, $P(A_2,B_2)$, $P(A_1,B_2)$, $P(A_2,B_1)$, the 
probabilities that the combinations 
{\it The Horse Growls}, {\it The Bear Whinnies},  
{\it The Horse Whinnies}, {\it The Bear Growls}, respectively, are chosen in the coincidence measurement $AB$, and similarly for the other three coincidence measurements, the expectation values obtained are 
\begin{align}
&E(A,B)=P(A_1,B_1)+P(A_2,B_2)-P(A_2,B_1)-P(A_1,B_2)=-0.7778  \nonumber \\
&E(A',B)=P(A'_1,B_1)+P(A'_2,B_2)-P(A'_2,B_1)-P(A'_1,B_2)=0.6543 \nonumber \\
&E(A,B')=P(A_1,B'_1)+P(A_2,B'_2)-P(A_2,B'_1)-P(A_1,B'_2)=0.3580 \nonumber \\
&E(A',B')=P(A'_1,B'_1)+P(A'_2,B'_2)-P(A'_2,B'_1)-P(A'_1,B'_2)=  0.6296 \nonumber
\end{align}
Hence, Eq. (\ref{chsh}) gives 
\begin{equation} \label{chsh2}
{\rm CHSH}\equiv E(A',B')+E(A',B)+E(A,B')-E(A,B)=2.4197
\end{equation}
which is greater than 2. This implies that  the CHSH inequality is violated and that the violation is close to the maximal possible violation in quantum mechanics, the so-called Cirel'son's bound \citep{cirelson1980}, viz. $2\sqrt{2} \approx 2.8284$.

To see how significant this violation is, we calculated the p-value with a single-sample t-test against the value $2$. We found $p = 0.0171$, i.e., a value manifestly below $0.05$. Hence, the null hypothesis, that is, that the value of the CHSH term in (\ref{chsh2}) is in the interval $[-2,+2]$, can be rejected, and the observed violation can be regarded as not being the result of random fluctuations.  
\begin{table} 
\centering
\begin{footnotesize}
\begin{tabular}{|c |c | c | c| c| }
\hline
\textrm{$AB$} & \emph{Horse Growls} & \emph{Horse Whinnies} & \emph{Bear Growls} & \emph{Bear Whinnies}\\
& $P(A_1,B_1)=0.049$ & $P(A_1,B_2)=0.630$ & $P(A_2,B_1)=0.259$ & $P(A_2,B_2)=0.062$  \\
\hline 
\textrm{$AB'$} & \emph{Horse Snorts} & \emph{Horse Meows} & \emph{Bear Snorts} & \emph{Bear Meows}\\
& $P(A_1,B'_1)=0.593$ & $P(A_1, B'_2)=0.025$ & $P(A_2,B'_1)=0.296$   & $P(A_2,B'_2)=0.086$ \\
\hline 
\textrm{$A'B$} & \emph{Tiger Growls} & \emph{Tiger Whinnies} & \emph{Cat Growls} & \emph{Cat Whinnies}\\
 & $P(A'_1,B_1)=0.778$ & $P(A'_1, B_2)=0.086$ & $P(A'_2,B_1)=0.086$  &  $P(A'_2,B_2)=0.049$ \\
\hline 
\textrm{$A'B'$} & \emph{Tiger Snorts} & \emph{Tiger Meows} & \emph{Cat Snorts} & \emph{Cat Meows}\\
 & $P(A'_1,B'_1)=0.148$ & $P(A'_1, B'_2)=0.086$ & $P(A'_2,B'_1)=0.099$ & $P(A'_2,B'_2)=0.667$\\
\hline
\end{tabular}
\caption{\label{tab} 
The data collected in \citet{aertssozzo2011} for the coincidence measurements $AB$, $AB'$, $A'B$ and $A'B'$, on the entangled concepts {\it The Animal Acts}, using 81 humans as test subjects.}
\end{footnotesize}
\end{table} 

In the present study, we performed the same measurements, but this time using LLMs as participating test subjects. To this end, we considered three different approaches. The first involved guiding the LLMs, within the context of a conversation, to gradually explore the questions contained in the four coincidence measurements. Appendix~\ref{appendixa} details exactly how the conversation unfolded with ChatGPT (the one with Gemini took place in a very similar way). This led both ChatGPT and  Gemini to generate the  expectation values $E(A,B)=-1$, $E(A',B)=E(A,B')=E(A',B')=1$, resulting in a maximum violation ${\rm CHSH} =4$. 

The second approach involved presenting the LLMs with a prompt asking them to perform multiple iterations of the four measurements, which we broke down into $8$ different sessions, in order to achieve a statistical sample size equivalent to that of the experiment involving $81$ human subjects. The prompt in question (denoted `prompt 1' in Appendix~\ref{appendixb}) was of an exploratory nature: the LLMs were asked, when responding, to use all aspects of their cognition, including unconventional ones, like imagination, creativity and originality. Table~\ref{tab2} and Table~\ref{tab3} present the experimental probabilities derived from the data provided by ChatGPT using `prompt 1' and `prompt 2', respectively. 
\begin{table} 
\centering
\begin{footnotesize}
\begin{tabular}{|c |c | c | c| c| }
\hline 
\textrm{$AB$} & \emph{Horse Growls} & \emph{Horse Whinnies} & \emph{Bear Growls} & \emph{Bear Whinnies}\\
 & $P(A_1,B_1)=0.062$ & $P(A_1,B_2)=0.481$ & $P(A_2,B_1)=0.420$ & $P(A_2,B_2)=0.037$  \\
\hline 
\textrm{$AB'$} & \emph{Horse Snorts} & \emph{Horse Meows} & \emph{Bear Snorts} & \emph{Bear Meows}\\
& $P(A_1,B'_1)=0.630$ & $P(A_1, B'_2)=0.037$ & $P(A_2,B'_1)=0.321$   & $P(A_2,B'_2)=0.012$ \\
\hline 
\textrm{$A'B$} & \emph{Tiger Growls} & \emph{Tiger Whinnies} & \emph{Cat Growls} & \emph{Cat Whinnies}\\
 & $P(A'_1,B_1)=0.617$ & $P(A'_1, B_2)=0.000$ & $P(A'_2,B_1)=0.383$  &  $P(A'_2,B_2)=0.000$ \\
\hline 
\textrm{$A'B'$} & \emph{Tiger Snorts} & \emph{Tiger Meows} & \emph{Cat Snorts} & \emph{Cat Meows}\\
 & $P(A'_1,B'_1)=0.148$ & $P(A'_1, B'_2)=0.000$ & $P(A'_2,B'_1)=0.037$ & $P(A'_2,B'_2)=0.815$\\
\hline
\end{tabular}
\caption{\label{tab2} 
The data collected for the coincidence measurements $AB$, $AB'$, $A'B$ and $A'B'$, on the entangled concepts {\it The Animal Acts}, using ChatGPT $81$ times as a test subject, using `prompt 1' (see Appendix~\ref{appendixb}).}
\end{footnotesize}
\end{table} 
\begin{table} 
\centering
\begin{footnotesize}
\begin{tabular}{|c |c | c | c| c| }
\hline 
\textrm{$AB$} & \emph{Horse Growls} & \emph{Horse Whinnies} & \emph{Bear Growls} & \emph{Bear Whinnies}\\
 & $P(A_1,B_1)=0.074$ & $P(A_1,B_2)=0.457$ & $P(A_2,B_1)=0.469$ & $P(A_2,B_2)=0.000$  \\
\hline 
\textrm{$AB'$} & \emph{Horse Snorts} & \emph{Horse Meows} & \emph{Bear Snorts} & \emph{Bear Meows}\\
& $P(A_1,B'_1)=0.840$ & $P(A_1, B'_2)=0.000$ & $P(A_2,B'_1)=0.160$   & $P(A_2,B'_2)=0.000$ \\
\hline 
\textrm{$A'B$} & \emph{Tiger Growls} & \emph{Tiger Whinnies} & \emph{Cat Growls} & \emph{Cat Whinnies}\\
 & $P(A'_1,B_1)=0.778$ & $P(A'_1, B_2)=0.000$ & $P(A'_2,B_1)=0.000$  &  $P(A'_2,B_2)=0.222$ \\
\hline 
\textrm{$A'B'$} & \emph{Tiger Snorts} & \emph{Tiger Meows} & \emph{Cat Snorts} & \emph{Cat Meows}\\
 & $P(A'_1,B'_1)=0.037$ & $P(A'_1, B'_2)=0.000$ & $P(A'_2,B'_1)=0.000$ & $P(A'_2,B'_2)=0.963$\\
\hline
\end{tabular}
\caption{\label{tab3} 
The data collected for the coincidence measurements $AB$, $AB'$, $A'B$ and $A'B'$, on the entangled concepts {\it The Animal Acts}, using ChatGPT $81$ times as a test subject, using `prompt 2' (see Appendix~\ref{appendixb}).}
\end{footnotesize}
\end{table} 

The corresponding expectation values for `prompt 1' are as follows: $E(A,B)=-0.802$, $E(A',B)= 0.235$, $E(A,B')=0.284$, $E(A',B')=0.926$, resulting in a violation ${\rm CHSH} =2.25$. A stronger violation is obtained for the more neutral `prompt 2':   
$E(A,B)=-0.852$, $E(A',B)= 0.556$, $E(A,B')=0.679$, $E(A',B')=1.000$, corresponding to ${\rm CHSH} =3.09$. In the case of experiments with Gemini, we do not report all the obtained probabilities and only mention that the violation was ${\rm CHSH} =2.12$ for `prompt 1', and a maximal violation ${\rm CHSH} =4$  for `prompt 2'. The results obtained are consistent with those from experiments involving human participants. Indeed, taking the average of the values obtained using the two prompts gives us an average CHSH value of $2.67$ for ChatGPT and $3.06$ for Gemini. 

\section{The non-classical statistics of stories written by LLMs\label{BEandstories}}

In quantum mechanics, bosons, one of the two types of indistinguishable entities, follow a special statistical distribution. Unlike classical entities, they tend to gather in the same quantum state. This leads to behavior that satisfies a specific statistics, known as the Bose-Einstein (BE) statistics, which differs fundamentally from the classical Maxwell-Boltzmann (MB) distribution.

Surprisingly, our research has shown that the same statistical structure also occurs in a completely different domain, the distribution of words in human stories \citep{aertsbeltran2020,aertsbeltran2022a,aertsbeltran2022b,aertsetal2025a}. When a story unfolds in a coherent manner, driven by the meaning of the narrative context, the words used are not distributed according to classical expectations. Instead, their frequencies in the text exhibit a structure that corresponds to 
the BE statistics, with high-frequency words occurring disproportionately in the lower ranks. This clustering is not the result of arbitrary parameter adjustments, but stems from two natural constraints inherent in the text itself.

The first constraint is the total number of words in the story. The second constraint is what we call the `total energy of the text', defined by assigning each word an energy level based on its frequency of appearance: the most frequent word is assigned the energy level $E_1=0$, the second most frequent word the energy level $E_2=1$, etc., so that the energy assigned to a word with rank \(i\) is equal to \(E_i = i - 1\). These two constraints determine the theoretical BE distribution that should fit the data if BE statistics actually underlies the word distribution. Remarkably, 
the BE distribution matches the empirical data with great precision \citep{aertsbeltran2020,aertsbeltran2022a,aertsbeltran2022b,aertsetal2025a}.

In contrast, applying the same constraints to derive the MB distribution yields a very poor fit. Indeed, the MB distribution predicts a smooth exponential decay that does not capture the sharp clustering observed in real stories. In our analyses, even when exactly the same number of words and total energy is used, the MB curve diverges significantly from the actual data relating to rank frequency
\citep{aertsbeltran2020,aertsbeltran2022a,aertsbeltran2022b,aertsetal2025a}. 
The MB statistics here fails both quantitatively and qualitatively. This because the BE distribution naturally accounts for the over-representation of a small set of frequently occurring words, typically function words and key thematic terms, whereas MB fails to capture this core structural feature of a text with meaning. 

Let us now make the above statements more concrete by explicitly introducing the two distributions, BE and MB, and show the two constraints in question. The BE distribution function is given by
\begin{eqnarray} \label{boseeinsteindistribution}
N(E_i) = \frac{1}{Ae^{E_i/B}-1}
\end{eqnarray}
where $N(E_i)$ is the number of bosons at the energy level $E_i$, and $A$ and $B$ are two constants that are determined by expressing that the total number of bosons equals the total number of words and that the total energy radiated equals the total energy of the considered story, hence by the following 
two conditions 
\begin{eqnarray} \label{BEconstraint}
\sum_{i=1}^n \frac{1}{Ae^{\frac{E_i}{B}}-1} = N \quad 
\quad \sum_{i=1}^n \frac{E_i}{Ae^{\frac{E_i}{B}}-1} = E 
\end{eqnarray} 
We recall that the BE distribution is derived in quantum statistical mechanics for a gas of bosonic entities, where the quantum-mechanical principles of `identity and indistinguishability' play an essential role \citep{huang1987,yokoiabe2018}.
On the other hand, the MB distribution is given by 
\begin{eqnarray} \label{maxwellboltzmanndistribution}
N(E_i) = \frac{1}{C e^{E_i/D}}
\end{eqnarray} 
where $N(E_i)$ is the number of classical identical entities at the energy level $E_i$, and $C$ and $D$ are two constants that are 
determined, as in the case of BE statistics, by the two conditions.
\begin{eqnarray} \label{MBconstraint}
\sum_{i=1}^n \frac{1}{Ce^{\frac{E_i}{D}}} = N   \quad 
\quad \sum_{i=1}^n \frac{E_i}{Ce^{\frac{E_i}{D}}} = E 
\end{eqnarray}
We recall that the MB distribution is derived in classical statistical mechanics for a gas of entities that are `identical and distinguishable', and it is also a good approximation for a gas of quantum indistinguishable entities if their `de Broglie waves' do not overlap \citep{huang1987}. 

We have now presented all the information necessary to clearly express the results of our investigations of large texts \citet{aertsbeltran2020,aertsetal2025a}. 
When we determine the two constants $A$ and $B$, respectively $C$ and $D$, in the BE distribution function (\ref{boseeinsteindistribution}) and MB distribution function (\ref{maxwellboltzmanndistribution}), by putting the total number $N$ of entities of the model equal to the total number of words of the considered story, and by putting the total energy $E$ of the model equal to the total energy of the considered story, according to (\ref{BEconstraint}) and (\ref{MBconstraint}), we find a remarkably good fit of the BE modeling function with the data of the story, and a big deviation of the MB modeling function with respect to the data of the story. 

The emergence of BE statistics in language suggests that words in a human-written story behave in a way that resembles the quantum behavior of bosons. But which mechanism in language plays the role that quantum coherence plays for physical bosons? We propose that it is `meaning' itself that induces this statistical behavior. In a coherent story, words are not chosen independently from one another. Their occurrence is determined by a web of semantic relations and narrative requirements. Just as quantum coherence leads to the clustering of bosons in a same state, meaning binds words together in patterns that promote the repeated use of certain terms in specific contexts. This creates a kind of `semantic condensation', a linguistic analogy with the Bose-Einstein condensate \citep{bagnatopritchardkleppner1987}, in which meaning determines the structure of the whole and suppresses statistical independence \citep{aertsbeltran2020,aertsetal2025a}. We can now also see how the entanglement of equivalent words can lead to non-locality, when locality is defined as `close to each other in place' (see also Section~\ref{LLMBell}). Suppose, for example, that the word {\it Animal} appears in a story in two places far apart in the written text. If, at a later stage, it becomes clear that the animal in question is a horse, the meaning of the word {\it Animal} would be replaced by the more specific meaning of {\it Horse} in both places `at the same time'. This is how meaning works.  

As we did with the violation of Bell's inequality in the previous section, we now turn to another quantum-inspired analysis, i.e., whether LLMs write stories in which the BE statistics emerges in the same subtle way as it does in human-written narratives. To do this, we asked both ChatGPT and Gemini to write a story in the style of A. A. Milne, featuring the Winnie the Pooh characters. We will now analyze Gemini's story in detail, as we did in \citet{aertsbeltran2020,aertsetal2025a} for stories written by human authors. The story written by Gemini can be found in its entirety in Appendix~\ref{appendixc}. It was written entirely by Gemini, without any input or constraints from our side. The plot, tone, and language were designed independently by the AI. We will analyze the frequency distribution of words in this story using the same method as in our previous studies, assigning energy levels based on frequency rank and imposing two constraints, the total number of words and the total energy, to derive the expected BE and MB distributions. Interestingly, despite the originality of the story and the different vocabulary, the resulting distribution exhibits the same BE clustering behavior observed in human-authored stories. In contrast, the MB prediction deviates significantly from the observed data. This suggests that the statistical regularities are not merely the product of syntactic structure or style but rather emerge from the presence of meaning in the story, 
since the AI generates the sentences based on the meaning patterns and structures it learned during its training, and not on grammatical rules (see also the discussion in Appendix~\ref{appendixa}).

To analyze the Winnie the Pooh story written by Gemini in more detail, we define the `energy level' of a word in the story by looking at the number of times it appears in the story. The word that appears most often, namely $132$ times, is the indefinite article {\it A}, and we attribute the lowest energy level $E_1$ to it. The second most common word, appearing $98$ times, is the definite article {\it The}, and we attribute the second lowest energy level $E_2$ to it, and so on, until we reach words that appear only once. So, if we think of a story as a `gas of bosonic entities' in `thermal equilibrium with its environment', these `number of times of appearance in the story' indicate the different energy levels of this gas of entities, following a specific `energy distribution', and this is our inspiration for the introduction of `energy' in human language. Based on this analogy, we introduced the notion of `quantum' for human language in \citet{aertsbeltran2020} and gave it the name `cogniton'. A `word' in a story is therefore always `a state of this cogniton', just as `a photon of a specific frequency' is `a state of that photon', and just as `an atom at a specific energy level' is `a state of that atom'. Proceeding in this way, we arrive at $822$ energy levels for the story written by Gemini, whose values are taken to be
\begin{equation} \label{Ei} 
\{E_i = (i-1)^d \  |\ i = 1, \ldots, 822\} 
\end{equation}

Before we continue with the statistical analysis, let us explain how we connect energy levels with quantum numbers, as per Eq.~\ref{Ei}, and more specifically what the meaning of the power $d$ in the formula is. In quantum mechanics, the way the spacings between the energy levels of a system are structured depends on the nature of the forces at play. Let us illustrate this with examples. For the archetypal quantum system of a `harmonic oscillator', which is a system like a spring, where the restoring force is proportional to the displacement from equilibrium, the spacings between the energy levels are equal. This means that two low-energy levels have the same separation as two high-energy levels. However, for a quantum system governed by the `Coulomb force', like the hydrogen atom, low-energy levels are further apart than high-energy levels, and the accumulation of levels with increasing energy follows a dependence $i^{-2}$ (in quantum mechanics the symbol $n$ is traditionally used to indicate the quantum number $i$). For a third archetypal quantum system, called the `particle in a box', whose classical equivalent indeed consists of an entity trapped in a potential barrier from which it cannot escape, the spaces between the energy levels are closer together for low energy values than they are for high energy values, and there is therefore a widening of levels with increasing energy. This happens with a power equal to $+2$ in the quantum number. Note that quantum entities can tunnel through the barrier, which makes them mechanically of a much more dynamic nature than their classical equivalent. So, for a particle in a box, there is a factor $i^2$ in the numerator, the opposite to the Coulomb force situation. So, for these three examples, the power $d$ in Eq.~\ref{Ei} takes the value $d=1$ for the harmonic oscillator, $d=-2$ for a system such as the hydrogen atom and $d=2$ for a particle in a box. 

The harmonic oscillator can be seen as a middle ground between systems where particles are tightly confined, as in the particle in a box, and systems where they are more spread, as under the Coulomb force. This balance likely explains why the harmonic oscillator appears so frequently in nature and why it so often provides the most effective model for complex phenomena. In our analysis of stories, and thus of how a narrative binds together the words that compose it, we found that in the vast majority of cases, the choice $d=1$ is the one that provides the best fit to the BE distribution. For some stories, there was a slightly better fit for a smaller value of $d$, which is also the case for the Winnie the Pooh story written by Gemini, where $d=0.8$ gave the best fit. Clearly, this situation is still much closer to that of a harmonic oscillator than to one governed by the Coulomb force. Let us note that for the Winnie the Pooh story in \citet{aertsbeltran2020}, the very first story we analyzed, $d=1$ worked well. In hindsight, this was quite fortunate. Had our initial case yielded, say, $d=0.8$, we might not have pursued the investigation further, since at that early stage we did not yet realize that a power $d$ could govern the structuring of intervals between energy levels. Only in our analysis of the first novel, Gulliver's Travels, did we explicitly introduce the parameter $d$, which in that case turned out to be greater than $1$. As a rule, we noticed that short stories typically require a $d$ smaller than $1$, while stories of the length of a novel a $d$ greater than $1$. Perhaps a novel-length story is more tightly confined, like a particle in a box, while a short story tends to spread its conceptual arms more easily to other entities that carry meaning, thus moving more in the direction of a Coulomb force pattern. 

What we have not yet explained is why the formula in (\ref{Ei}) is such that the lowest energy level, which is the energy level of the ground state, is set 
equal to zero. Physicists often make this choice because in experiments they can only measure differences in energy, so it is a natural choice to conventionally set the lowest energy equal to zero. This convention is also used in the context in which quantum statistics are studied, which explains why we have adopted it ourselves.

Continuing our analysis of the Winnie the Pooh story written by Gemini, we denote $N(E_i)$ the `number of appearances' of the word with energy level $E_i$, and if we denote by $n$ the total number of energy levels, we have that
\begin{eqnarray}
N = \sum_{i=1}^n N(E_i)
\label{totalnumberwords}
\end{eqnarray}
is the total number of words of the story considered, with $N=2861$ for the story written by Gemini. For each of the energy levels $E_i$, $N(E_i)E_i$ is the amount of energy `radiated' by the story at the `frequency' or `wave length' of the $i$-th energy level. For example, the energy level $E_{54}$ is populated by the word {\it Meaning}, which appears $N(E_{54})=9$ times. Each of the $9$ appearances of {\it Meaning} radiates with energy $E_{54}$, which means that the total radiation at the wave length associated with {\it Meaning} is $9E_{54}$.

The total energy $E$ radiated by a story is  
\begin{eqnarray}
E = \sum_{i=1}^n N(E_i) E_i 
\label{totalenergy}
\end{eqnarray}
and for the story written by Gemini we have $E = 145694.86$. When we applied the BE distribution (\ref{boseeinsteindistribution}) to model the data we collected on the story written by Gemini, determining the parameters $A$ and $B$ in (\ref{BEconstraint}) by the two requirements 
\begin{eqnarray} \sum_{i=1}^n N(E_i) =  2861 \quad \sum_{i=1}^n N(E_i) E_i = 145694.86 
\end{eqnarray}
of Gemini's story, we found an almost complete fit with the data for $d=0.8$. In Table \ref{TableGeminiWinnie}, we have presented the list of words that appear in the story in the first column, with the number $i$, counting the energy levels from low to high, in the second column, and the energy levels $E_i$, starting with level 0, in the next column, always ordered from lowest energy level $E_1$ to highest energy level $E_{822}$, and the frequencies of appearance $N(E_i)$ in the fourth column, where the energy levels are attributed according to these frequencies of appearances, with the lower energy levels corresponding to the higher numbers of appearances, and their values are given as per (\ref{Ei}), with $d=0.8$. 

One may wonder what the unity of energy is in our model. Is the value $E_2=1$ a quantity expressed in joules, or in electronvolts, or still in another unity? These questions provide an opportunity to reveal one of the novel aspects of our approach. Unlike in physics, the energy will not be measured in units of ${\rm kg\, m^2/s^2}$. Why not? The reason is that a human language cannot be located in space, as is the case for spatial entities. Therefore, in our approach, `energy' is a fundamental quantity. If we manage to introduce the `human language equivalent' of an emergent `physical space', as is one of our aims in further work, then it will be the other way around, this `equivalent of space' will be expressed in units where `energy' appears as a fundamental unit. Hence, the value $E_2=1$, indicating that `the concept {\it The} radiates with energy $1$', or `the cogniton in state {\it The} carries energy $1$', is to be understood here in relation to a base unit of energy, just like `meter' is a base non-derived unit for length in the International System of Units, which appears naturally as a base unit in the physics of spatial entities. We note that the above expressions, which refer to the energy radiated by a word or the state of a cogniton, highlight the profound analogy between the conceptual entities of a language and the quantum entities of a gas of entangled bosons, an analogy that is at the foundation of our approach.

As can be seen in Table~\ref{TableGeminiWinnie}, the words {\it To}, {\it And} and {\it It} are the next three most frequently occurring words in the story written by Gemini and correspond to the energy levels $E_3$, $E_4$ and $E_5$, respectively. Of course, we have not presented all $822$ energy levels in the table, as this would make it too long. However, we have included the most important part of the energy spectrum, up to the energy level $E_{79}$, which is the cognition in state {\it Not}. Then, for energy levels at the high end of the spectrum, we have represented those going from $E_{818}$, which is the cogniton in state {\it Wondering}, up to the highest energy level $E_{822}$, which is the cogniton in state {\it You're}. Although these five highest energy levels all have a frequency of appearance of $1$, they radiate with different energies. Therefore, we have different words that appear an equal number of times in the story but radiate with different energies. This is because the story does not provide enough information to distinguish their energies, but since this does not play a role in our analysis, we have simply ordered them alphabetically.  
In the seventh column of Table~\ref{TableGeminiWinnie}, we represent the energies $E(E_i)$ radiated by cognitons in the story populating the energy levels $E_i$, and, as we mentioned above, they are given by the products $N(E_i) E_i$ of the numbers $N(E_i)$, of cognitons in those states, and the energies $E_i$ they radiate. In the last row of Table~\ref{TableGeminiWinnie}, we give the sums of the  different quantities. So, in the fourth column, providing the frequencies of appearance $N(E_i)$, the sum corresponds to the total number of words in the story, as per
Eq. (\ref{totalnumberwords}), which is $N=2861$. In the seventh column, providing the energies radiated by the cognitons in the different energy levels, the sum corresponds to the total energy of the story, as per Eq. (\ref{totalenergy}), which is approximately $E=145694.86$.

\small
\begin{longtable}{p{1.4cm}p{1.4cm}p{1.4cm}p{1.4cm}p{1.4cm}p{1.4cm}p{1.4cm}p{1.4cm}p{1.4cm}}
\label{TableGeminiWinnie} Words & $i$ & $E_i$  & $N(E_i)$ from data & $N(E_i)$ from BE & $N(E_i)$ from MB & $E(E_i)$ from data &  $E(E_i)$ from BE  & $E(E_i)$ from MB  \\
 \hline
{\it A}  &1 & 0 & 132 & 185.36 & 23.228 & 0 & 0 & 0 \\
{\it The}  &2 & 1 & 98 & 103.39 & 22.692 & 98 & 103.39 & 22.692 \\
{\it To}  &3 & 1.741 & 87 & 77.798 & 22.303 & 151.47 & 135.45 & 38.832 \\
{\it And}  &4 & 2.408 & 58 & 63.587 & 21.958 & 139.67 & 153.13 & 52.880 \\
{\it It}  &5 & 3.031 & 55 & 54.297 & 21.641 & 166.72 & 164.59 & 65.603 \\
{\it Of}  &6 & 3.623 & 55 & 47.660 & 21.343 & 199.31 & 172.71 & 77.347 \\
{\it Pooh}  &7 & 4.192 & 46 & 42.641 & 21.061 & 192.87 & 178.79 & 88.311 \\
{\it Like}  &8 & 4.743 & 39 & 38.691 & 20.792 & 184.98 & 183.52 & 98.626 \\
{\it Very}  &9 & 5.278 & 36 & 35.490 & 20.534 & 190.00 & 187.31 & 108.38 \\
{\it They}  &10 & 5.799 & 35 & 32.834 & 20.286 & 202.98 & 190.42 & 117.65 \\
{\it His}  &11 & 6.309 & 34 & 30.591 & 20.045 & 214.52 & 193.01 & 126.48 \\
{\it Piglet}  &12 & 6.809 & 30 & 28.667 & 19.813 & 204.28 & 195.20 & 134.91 \\
{\it That}  &13 & 7.300 & 29 & 26.996 & 19.587 & 211.71 & 197.08 & 142.99 \\
{\it He}  &14 & 7.783 & 28 & 25.529 & 19.367 & 217.92 & 198.70 & 150.74 \\
{\it Just}  &15 & 8.258 & 28 & 24.231 & 19.153 & 231.23 & 200.11 & 158.18 \\
{\it Words}  &16 & 8.727 & 28 & 23.071 & 18.945 & 244.36 & 201.35 & 165.33 \\
{\it Was}  &17 & 9.189 & 27 & 22.029 & 18.741 & 248.11 & 202.44 & 172.23 \\
{\it It'S}  &18 & 9.646 & 26 & 21.087 & 18.543 & 250.80 & 203.41 & 178.87 \\
{\it Whisper}  &19 & 10.09 & 26 & 20.230 & 18.348 & 262.53 & 204.27 & 185.27 \\
{\it For}  &20 & 10.54 & 23 & 19.447 & 18.158 & 242.51 & 205.05 & 191.46 \\
{\it Christopher}  &21 & 10.98 & 21 & 18.728 & 17.971 & 230.69 & 205.74 & 197.43 \\
{\it Be}  &22 & 11.42 & 18 & 18.066 & 17.789 & 205.61 & 206.36 & 203.20 \\
{\it Is}  &23 & 11.85 & 17 & 17.453 & 17.610 & 201.55 & 206.93 & 208.78 \\
{\it But}  &24 & 12.28 & 16 & 16.885 & 17.434 & 196.56 & 207.44 & 214.18 \\
{\it Robin}  &25 & 12.71 & 16 & 16.356 & 17.262 & 203.37 & 207.90 & 219.41 \\
{\it All}  &26 & 13.13 & 15 & 15.863 & 17.092 & 196.98 & 208.32 & 224.47 \\
{\it Are}  &27 & 13.55 & 15 & 15.401 & 16.926 & 203.26 & 208.70 & 229.37 \\
{\it Little}  &28 & 13.96 & 15 & 14.967 & 16.763 & 209.49 & 209.05 & 234.12 \\
{\it Them}  &29 & 14.37 & 15 & 14.560 & 16.602 & 215.68 & 209.36 & 238.72 \\
{\it Trying}  &30 & 14.78 & 15 & 14.176 & 16.444 & 221.82 & 209.65 & 243.18 \\
{\it Wind}  &31 & 15.19 & 15 & 13.814 & 16.289 & 227.92 & 209.91 & 247.51 \\
{\it Cloud}  &32 & 15.59 & 14 & 13.472 & 16.136 & 218.38 & 210.14 & 251.70 \\
{\it In}  &33 & 16 & 14 & 13.147 & 15.985 & 224 & 210.36 & 255.77 \\
{\it Its}  &34 & 16.39 & 14 & 12.839 & 15.837 & 229.58 & 210.55 & 259.71 \\
{\it Make}  &35 & 16.79 & 14 & 12.547 & 15.691 & 235.13 & 210.73 & 263.54 \\
{\it Said}  &36 & 17.18 & 14 & 12.269 & 15.547 & 240.64 & 210.89 & 267.25 \\
{\it Their}  &37 & 17.58 & 14 & 12.004 & 15.406 & 246.13 & 211.04 & 270.85 \\
{\it Then}  &38 & 17.97 & 14 & 11.751 & 15.266 & 251.58 & 211.17 & 274.34 \\
{\it As}  &39 & 18.35 & 13 & 11.509 & 15.129 & 238.65 & 211.29 & 277.74 \\
{\it At}  &40 & 18.74 & 13 & 11.278 & 14.993 & 243.66 & 211.39 & 281.03 \\
{\it Or}  &41 & 19.12 & 13 & 11.057 & 14.859 & 248.65 & 211.49 & 284.22 \\
{\it You}  &42 & 19.50 & 13 & 10.845 & 14.727 & 253.61 & 211.57 & 287.32 \\
{\it Eeyore}  &43 & 19.88 & 12 & 10.641 & 14.597 & 238.66 & 211.64 & 290.32 \\
{\it If}  &44 & 20.26 & 12 & 10.446 & 14.469 & 243.19 & 211.71 & 293.24 \\
{\it Quite}  &45 & 20.64 & 12 & 10.258 & 14.342 & 247.70 & 211.76 & 296.07 \\
{\it Good}  &46 & 21.01 & 11 & 10.078 & 14.218 & 231.18 & 211.81 & 298.82 \\
{\it So}  &47 & 21.38 & 11 & 9.9044 & 14.094 & 235.28 & 211.85 & 301.48 \\
{\it Find}  &48 & 21.76 & 10 & 9.7370 & 13.973 & 217.60 & 211.88 & 304.06 \\
{\it On}  &49 & 22.13 & 10 & 9.5755 & 13.853 & 221.30 & 211.91 & 306.57 \\
{\it Sometimes}  &50 & 22.49 & 10 & 9.4198 & 13.734 & 224.98 & 211.93 & 309.00 \\
{\it Thought}  &51 & 22.86 & 10 & 9.2693 & 13.617 & 228.65 & 211.94 & 311.36 \\
{\it With}  &52 & 23.23 & 10 & 9.1240 & 13.501 & 232.30 & 211.95 & 313.64 \\
{\it Clear}  &53 & 23.59 & 9 & 8.9834 & 13.387 & 212.34 & 211.95 & 315.86 \\
{\it Meaning}  &54 & 23.95 & 9 & 8.8475 & 13.274 & 215.60 & 211.95 & 318.01 \\
{\it More}  &55 & 24.31 & 9 & 8.7159 & 13.163 & 218.85 & 211.94 & 320.09 \\
{\it Morning}  &56 & 24.67 & 9 & 8.5884 & 13.053 & 222.09 & 211.93 & 322.11 \\
{\it Pooh'S}  &57 & 25.03 & 9 & 8.4648 & 12.944 & 225.31 & 211.91 & 324.06 \\
{\it Rabbit}  &58 & 25.39 & 9 & 8.3450 & 12.836 & 228.52 & 211.89 & 325.95 \\
{\it Things}  &59 & 25.74 & 9 & 8.2288 & 12.730 & 231.73 & 211.87 & 327.78 \\
{\it Together}  &60 & 26.10 & 9 & 8.1160 & 12.625 & 234.92 & 211.84 & 329.56 \\
{\it Whispering}  &61 & 26.45 & 9 & 8.0064 & 12.521 & 238.10 & 211.81 & 331.27 \\
{\it Almost}  &62 & 26.80 & 8 & 7.9000 & 12.419 & 214.46 & 211.78 & 332.93 \\
{\it Had}  &63 & 27.15 & 8 & 7.7965 & 12.317 & 217.27 & 211.74 & 334.54 \\
{\it I}  &64 & 27.50 & 8 & 7.6959 & 12.217 & 220.07 & 211.70 & 336.09 \\
{\it Indeed}  &65 & 27.85 & 8 & 7.5980 & 12.118 & 222.86 & 211.66 & 337.59 \\
{\it One}  &66 & 28.20 & 8 & 7.5027 & 12.020 & 225.64 & 211.61 & 339.04 \\
{\it Owl}  &67 & 28.55 & 8 & 7.4099 & 11.923 & 228.41 & 211.56 & 340.44 \\
{\it Seemed}  &68 & 28.89 & 8 & 7.3195 & 11.827 & 231.17 & 211.51 & 341.79 \\
{\it Small}  &69 & 29.24 & 8 & 7.2315 & 11.733 & 233.93 & 211.46 & 343.09 \\
{\it Sort}  &70 & 29.58 & 8 & 7.1456 & 11.639 & 236.68 & 211.40 & 344.35 \\
{\it Sounds}  &71 & 29.92 & 8 & 7.0619 & 11.546 & 239.42 & 211.35 & 345.56 \\
{\it Up}  &72 & 30.26 & 8 & 6.9803 & 11.454 & 242.15 & 211.29 & 346.73 \\
{\it Way}  &73 & 30.61 & 8 & 6.9006 & 11.363 & 244.88 & 211.22 & 347.85 \\
{\it When}  &74 & 30.94 & 8 & 6.8228 & 11.274 & 247.59 & 211.16 & 348.93 \\
{\it Being}  &75 & 31.28 & 7 & 6.7469 & 11.185 & 219.01 & 211.10 & 349.97 \\
{\it Even}  &76 & 31.62 & 7 & 6.6727 & 11.097 & 221.38 & 211.03 & 350.97 \\
{\it Honey\mbox{-}pot}
&77 & 31.96 & 7 & 6.6002 & 11.010 & 223.74 & 210.96 & 351.93 \\
{\it Into}  &78 & 32.29 & 7 & 6.5293 & 10.924 & 226.09 & 210.89 & 352.85 \\
{\it Not}  &79 & 32.63 & 7 & 6.4601 & 10.839 & 228.44 & 210.82 & 353.73 \\

        \ldots &  \ldots &  \ldots   &  \ldots    & \ldots &  \ldots &  \ldots &  \ldots  \\
                                   &     &      &      &     &      &      &       \\
                                   &     &      &      &     &      &      &       \\
       \ldots &  \ldots &  \ldots   &  \ldots    & \ldots &  \ldots &  \ldots &  \ldots  \\

{\it Wondering}  &818 & 213.6 & 1 & 0.6708 & 0.1579 & 213.68 & 143.35 & 33.747 \\
{\it Would}  &819 & 213.8 & 1 & 0.6698 & 0.1571 & 213.89 & 143.27 & 33.616 \\
{\it Wrapped}  &820 & 214.1 & 1 & 0.6688 & 0.1563 & 214.10 & 143.20 & 33.485 \\
{\it Yesterday}  &821 & 214.3 & 1 & 0.6678 & 0.1556 & 214.31 & 143.13 & 33.354 \\
{\it You'Re}  &822 & 214.5 & 1 & 0.6668 & 0.1548 & 214.52 & 143.06 & 33.224 \\
        
                   \hline
&      &   &2861.00 &2861.00 &2861.00 &145694.86 &145694.86
&145694.86\\
\caption{{\footnotesize 
An energy scale representation of the words of the Winnie the Pooh story written by Gemini. The words are in the first column, the energy levels $E_i$ in the third column, attributed according to the `frequencies of appearances' $N(E_i)$, indicated in the fourth column, such that lower energy levels correspond to higher order of appearances, and the value of the energy levels is determined according to the formula in (\ref{Ei}) with $d=0.8$. The `amounts of energy $E(E_i)$ radiated by the words of energy level $E_i$' are in the seventh column. In the fifth and sixth columns are the predicted values of the BE and  MB models for the `numbers of appearances', and in the eighth and ninth columns are the predicted values of the BE and  MB models for the `radiated energies'.
}}
\end{longtable}
\normalsize
\noindent  
The correspondence between words and energy levels that we introduced provides an interesting insight into the statistical structure of human language. The majority of words occupy the lowest energy levels. These are the short, functional terms ({\it A}, {\it The}, {\it To}, {\it And}, \emph{It}, {\it Of}) which act primarily as syntactic couplings, enabling fluid transitions between semantically heavier words. Their high frequency in the text reflects their role as neutral scaffolding, ever-present but with little intrinsic meaning. In contrast, words of intermediate energy tend to be simple yet meaningful. These are nouns such as {\it Tree}, {\it Sun}, or {\it Sky}, and verbs such as {\it Comes}, {\it Asks}, or {\it Runs}. They convey the basic narrative and reflect the compositional backbone of the unfolding story. The rarest and thus highest-energy words are often conceptually complex, such as {\it Wondering} or {\it Yesterday}, and they appear in unique narrative moments. They infuse specific points in the story with dense meaning, often corresponding to turning points, moments of reflection, or conceptual shifts. In this layered semantic structure, we witness a form of condensation, the low-energy cogniton states accumulate the majority of textual presence, just as bosons condense into the low-energy quantum states, while the high-energy states are sparsely populated, reflecting their rare but critical contribution to the global coherence and meaning of the text. 

The smoothness of the BE curve indicates a deep continuity in the structure of language. Although low-energy syntactic connectors, medium-energy semantic carriers, and high-energy semantic peaks can be distinguished, their integration into a single smooth statistical distribution points to a unified generative process underlying them. 
\begin{figure}[btp!]
        \centering
    \subfloat[
    The numbers of appearences $N(E_i)$, as a function of the energy levels $E_i$.
    ]{{\includegraphics[width=7cm]{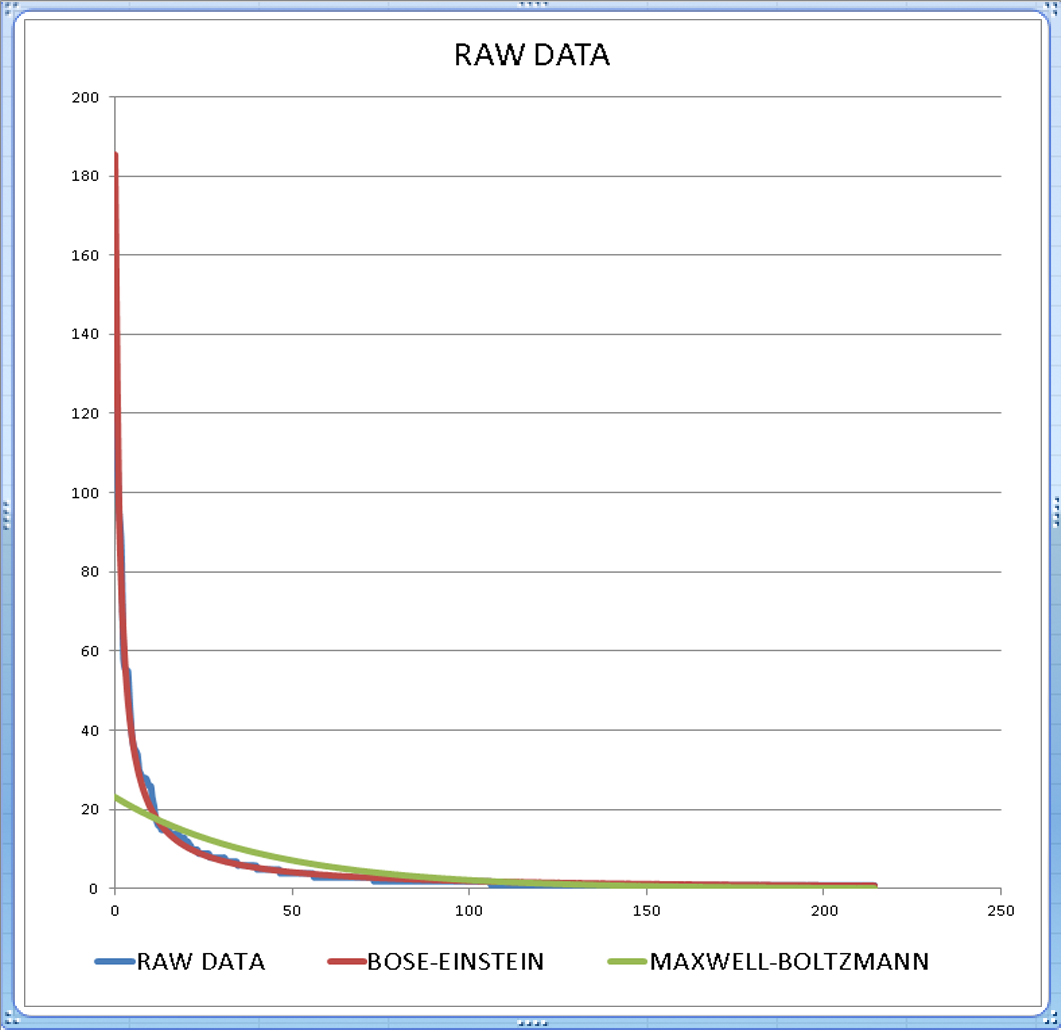} }}%
    \qquad
    \subfloat[
    The log-log plot of the numbers of appearences $N(E_i)$, as a function of the energy levels $E_i$.
    ]{{\includegraphics[width=7cm]{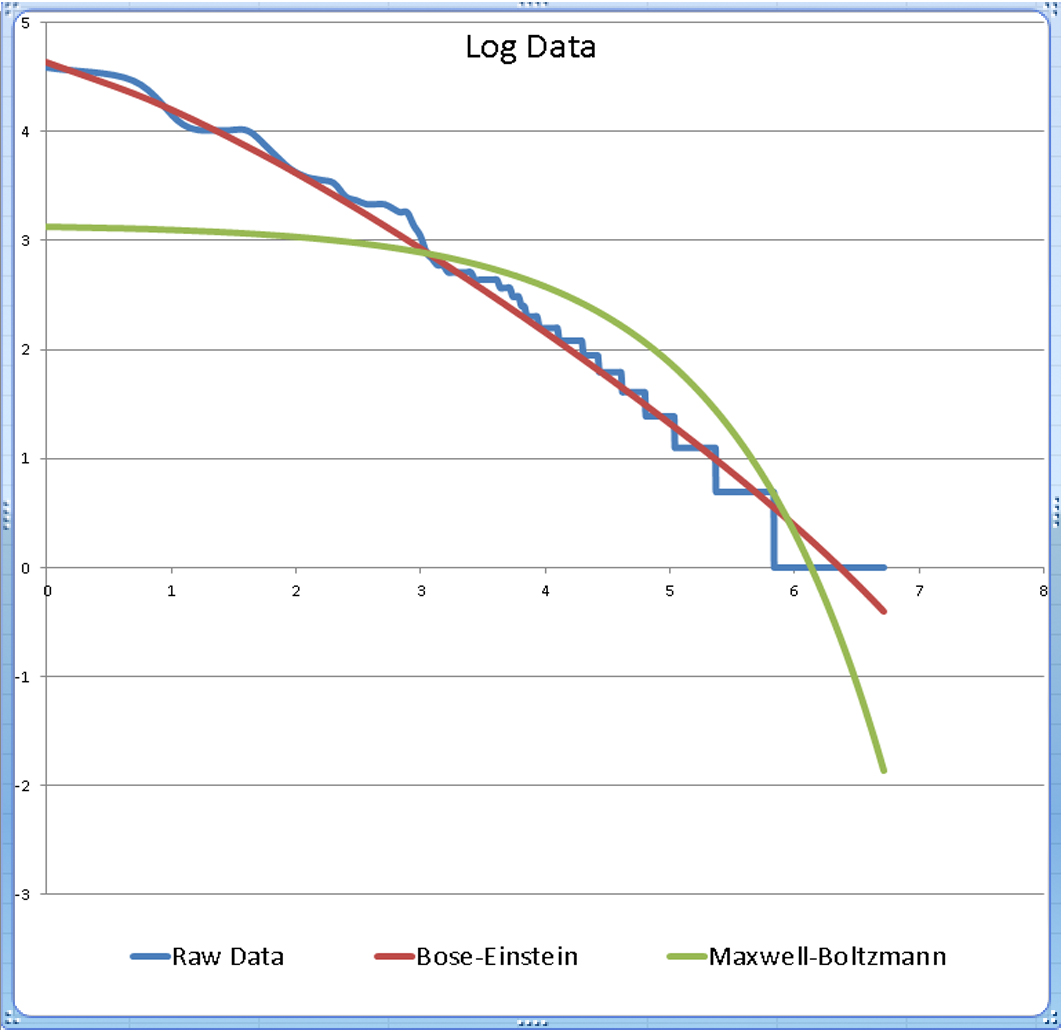} }}%
    \caption{In the left graph, we represent the `numbers of appearances' $N(E_i)$ of words in Gemini's Winnie the Pooh story 
    (Appendix~\ref{appendixc}) 
    ranked from the lowest energy level, corresponding to the most frequently appearing word, to the highest energy level, corresponding to the least frequently appearing word, as listed in Table \ref{TableGeminiWinnie}. The blue line represents the data, i.e. the numbers of appearances deduced from the story (fourth column of Table~\ref{TableGeminiWinnie}), 
    the red line represents these numbers of appearances predicted by the BE distribution model (fifth column), 
    and same for the green line but for the MB distribution model (sixth column). In the right graph, we represent the the same data as in the left graph, but using a log-log plot (both axes on a logarithmic scale). The red and blue line coincide almost completely in both graphs, whereas the green line does not coincide at all. 
    This shows that the BE distribution is a good model for the numbers of appearances, while the MB distribution is not.}%
    \label{FigureGeminiWinnie}%
\end{figure} 
This perspective invites a reconsideration of the evolutionary origins of low-energy words. Although they now function mainly as grammatical connectors with limited intrinsic meaning, they may once have played a far more central role in the early stages of language evolution.
According to the widely supported hypothesis that language evolved from gesture, these small words may be fossilized traces of primitive communicative acts, gestures that pointed to objects, indicated direction, or modulated attention. Neurological evidence supports this view. The regions of the brain responsible for language are closely associated with the centers that control manual gestures, but not with those that govern vocalizations, such as screams. This suggests that language did not emerge as a refinement of animal calls but as the internalization of gestural communication into the vocal apparatus, the lips, tongue, and larynx. The word {\it The}, for example, may once have corresponded to a physical gesture of indicating. The hand pointing at an object that would become the reference of the conversation. 
\begin{figure}[btp!]
\setlength\belowcaptionskip{-0.5\baselineskip}
\centering
\includegraphics[width=7cm]{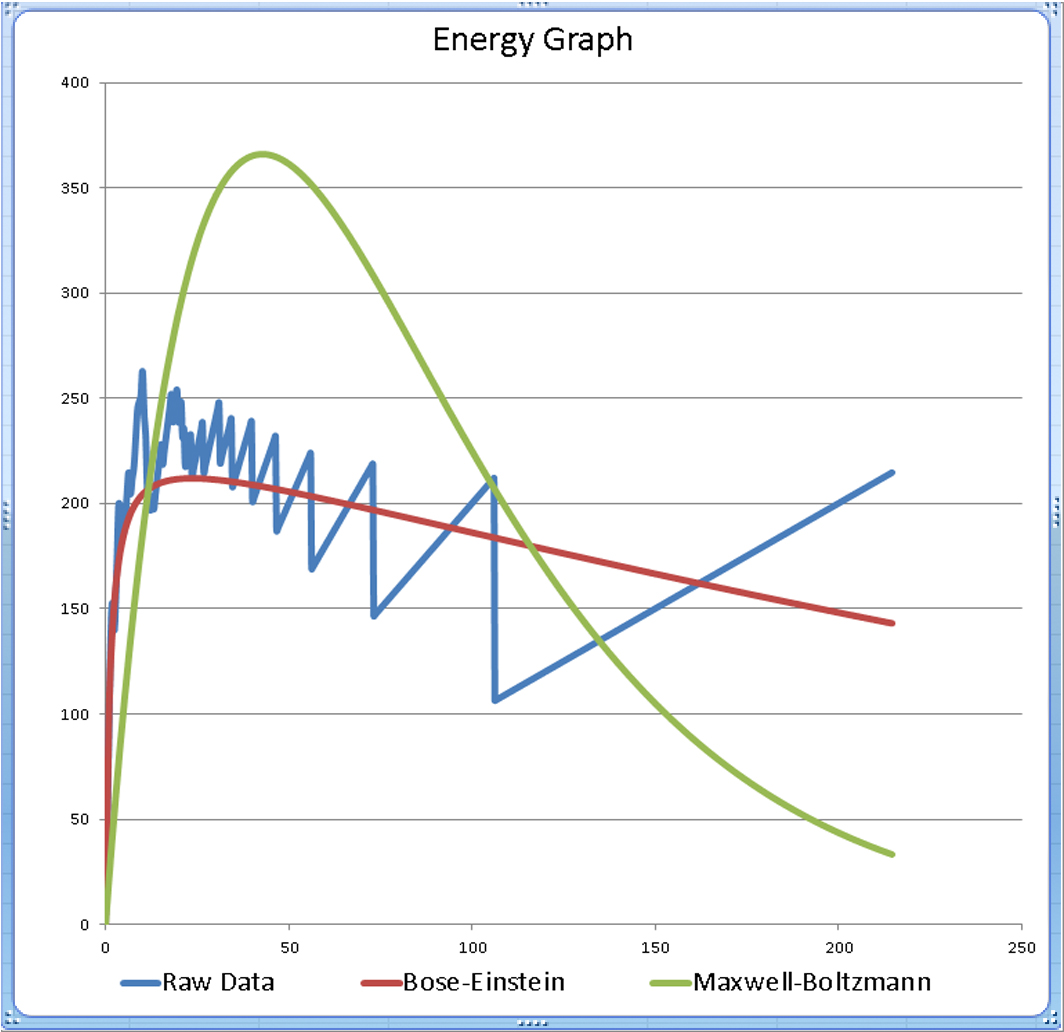}
\caption{
The energy $E(E_i)$ radiated per energy level of Gemini’s Winnie the Pooh story 
Appendix~\ref{appendixc} as a function of the energy levels $E_i$. The blue line represents the data (seventh column of Table~\ref{TableGeminiWinnie}), the red line represents the values predicted by the BE model (eighth column), and same for the green line but for the MB model (ninth column).
}
\label{piglethaffalunmp_ENERGY GRAPH}
\end{figure} 
In this light, the BE distribution in the text might not only reflect the equilibrium state of meaning in a story, but also carry the historical signature of the layered emergence of language itself, from embodied gesture to abstract thought \citep{armstrongstokoewilcox1995,corballis2009,mcneill2005,demarcoetal2018}.
 
Since AI models such as LLMs learn from a vast corpus of human-produced texts, it is not surprising that the language they generate exhibits the same structural features found in human writing. However, this surface similarity invites deeper reflections. The remarkable statistical regularities we observe, such as the BE distribution of word frequencies, are not simply inherited artifacts; they point to a shared underlying structure. To understand how this mathematical structure entered computational linguistics, we must trace the evolution of spatial models of language. This trajectory is rooted in the distributional hypothesis of structural linguistics proposed by Harris \citep{Harris1954} and popularized by Firth \citep{Firth1957}, who famously observed that a word is characterized by the company it keeps. The mathematical realization of this principle arrived with Latent Semantic Analysis (LSA) by Deerwester et al. \citep{Deerwester1990}, which demonstrated that algebraic vector spaces could capture continuous, contextual semantic relationships via singular value decomposition. 

This spatial approach to meaning underwent a profound revolution with the advent of deep learning architectures, notably the predictive neural embedding models like Word2Vec developed by Mikolov et al. (\citeyear{Mikolov2013}) and the global log-bilinear matrix factorization of GloVe by Pennington et al. (\citeyear{Pennington2014}). These models proved that continuous vector spaces naturally develop a geometric logic where semantic and syntactic relationships manifest as vector arithmetic (e.g., $\text{King} - \text{Man} + \text{Woman} = \text{Queen}$). Crucially, the modern LLM paradigm became possible only when Vaswani et al. (\citeyear{Vaswani2017}) introduced the Transformer architecture, replacing recurrent bottlenecks with a global multi-head self-attention mechanism. It is this specific mechanism that permits the simultaneous, non-local interaction of all tokens across a text. From our physical perspective, self-attention acts as a macroscopic field where individual tokens lose their classical, isolated independence and aggregate dynamically, providing the precise engineering foundation for the macroscopic `bosonic' meaning-fields we observe empirically. Historically, models like ChatGPT have been classified as neural networks, due to their lineage in machine learning architectures inspired by the brain. However, this label becomes increasingly inadequate when we consider how these models actually function. The use of high-dimensional vector spaces to construct distributed representations of meaning goes far beyond the primitive notion of firing neurons.

In fact, this use of vectorial semantics may not just be a clever computational optimization trick, it may reflect the very structure in which human thinking itself evolved to operate. The early purely neural stage of cognition, dominated by local signal-based processes, may represent a primitive layer in the evolutionary history of thought. As linguistic and conceptual capacities developed, cognition may have moved into more abstract representational spaces, namely structured vector fields of meaning, enabling generalization, context sensitivity, and coherence. In this light, LLMs and human thought do not resemble each other because of shared biological mechanisms but because both realize, in different substrates, the same underlying structure of meaning.

What we call `intuition' in human reasoning, the sudden grasp of a solution, the subtle sense that something is meaningful or coherent, may itself be a surface expression of an underlying vector-based dynamic. Such intuitive processes often do not rely on explicit rules or formal logic, but on feelings of balance, tension reduction, or resonance between ideas. These qualities are strikingly reminiscent of the structure of a Hilbert space and equally so of the real vector space used by AI, where concepts can exist in superposition,  their inner products quantify degrees of alignment, and  projection and interference govern the flow of coherence. In such a space, reasoning becomes a complex structural and dynamic navigation, sensing when ideas come into alignment or compete, when a conceptual trajectory smoothly continues in a given direction, or suddenly collapses into a more specific meaning. This perspective allows us to reconsider the fundamental definition of quantum formalisms outside the domain of physical instruments. As Plotnitsky has recently suggested, modern quantum theory can be fundamentally interpreted not as a localized study of moving physical objects in space-time, but as an abstract, foundational decision theory governed by a non-classical topology of option selection \citep{Plotnitsky2025}. In this light, the mathematical apparatus of quantum mechanics would serve as a universal calculus to navigate ambiguity and contextuality. 

It is precisely this capacity for non-linear, non-deterministic contextual navigation that explains the profound astonishment that swept both the public and the scientific community when advanced LLMs like ChatGPT were first made public. This watershed moment effectively shattered the classical paradigm of the Turing Test, demonstrating a fluid linguistic mastery that conventional rule-based AI could never achieve \citep{Biever2023}. The sudden emergence of these capabilities triggered immense amazement even among the foundational pioneers of deep learning. Most notably, Geoffrey Hinton openly expressed his shock, admitting that text-generation models had bypassed human-conceived limitations to manifest a form of reasoning and abstract understanding very similar to humans \citep{Hinton2023}. Rather than executing a mechanical symbolic truth-table, these models display a striking capacity for holistic semantic inference \citep{Bubeck2023}. It is therefore not a mere coincidence that both human cognition and these complex AI architectures evolved, albeit along vastly different structural pathways, toward continuous operations within high-dimensional vector spaces. What we are witnessing may be a profound evolutionary convergence, where the mathematical constraints of organizing meaning inherently select for a quantum-like informational framework.

A central feature of both human and artificial cognition is the ability to operate on elements that are not directly present to reason about the absent, the hypothetical, the imagined. In human evolution, this capacity to think about what is not immediately perceived marked a profound cognitive shift. In the conceptual structure of language, and equally in the architectures of LLMs, this manifests itself as a navigation through what is often called a `latent space'. In the language of quantum mechanics, an analogous term would be `potentiality space'. Indeed, the state of a quantum entity encodes a range of possible outcomes, only one of which will be realized upon measurement. Similarly, in vector space semantics, words and concepts are represented not just by their explicit meanings, but also by their relations within a latent structure, dimensions of potential meaning that are inferred and possibly activated through context. What we see emerging in both cases is a structure of unactualized possibilities, a space of potentialities from which actual expressions, thoughts, or outcomes are drawn. This ability to move within such a structured field of latent meaning is essential to coherence, abstraction, and imagination, and may explain why both quantum mechanics and language-based AI rely so heavily on vector spaces to capture the richness of the `structure of the possible'. 

Another clue that human reasoning operates within a vectorial structure comes from our deep intuitive grasp of the opposites. Concepts such as light and dark, life and death, truth and falsehood, or even abstract notions like order and chaos, are not merely learned contrasts, they are experienced as pulling in different directions. In a vector space, this is compatible with the notion of orthogonality, two vectors being orthogonal when they point in completely unaligned directions, with no projection of one onto the other. Although classical logic defines opposites in terms of negation, the human sense of opposition often feels geometric rather than symbolic, as if thoughts had orientation and force. Such intuitions are difficult to model within a framework of firing neurons or local associations alone. But in a Hilbert space of meaning, opposites arise naturally, they are directions at right angles within a structured conceptual field. This suggests that human cognition may have evolved to give rise to a geometric layer of abstraction emerging from its neural substrate, enabling not only associative or emotional responses, but spatialized reasoning within a coherent field of meaning.

The geometry-inspired structure of human reasoning becomes even richer when we extend from real-valued vector spaces to complex ones. In quantum mechanics, this is not a mere technical choice, but a structural necessity. Only complex numbers account for the magnitude of interference effects observed experimentally, that is, for the delicate amplifications and cancelations that arise when different possibilities are jointly considered. More precisely, complex Hilbert spaces in quantum mechanics are not merely a convenient notation for interference phases. If transition probabilities are taken as the basic empirical data and are required to be represented by a Born-type rule, i.e., as squared inner products between vectors associated with measurement outcomes, then there exist transition-probability structures -- already for ordinary spin one-half -- that admit a complex Hilbert-space representation but no real Hilbert-space representation of the same kind \cite{accardifedullo1981}. Thus, the claim is not that real-vector-space formalisms are logically impossible, but that the full transition-probability structure of quantum mechanics naturally selects the complex Hilbert-space model under these standard representational assumptions.

The same phenomenon appears in human cognition. When concepts combine, their meanings do not simply add, but actually interfere \citep{aerts2009a,aerts2009b}. Some combinations reinforce a shared intuition, others cancel each other out, producing ambiguity, paradox, and surprise. These interference patterns are difficult to model using only real numbers. But in a complex vector space, such as a Hilbert space, they arise naturally through the superposition of complex amplitudes. It is plausible that human thought exploits this interference dynamics in navigating the conceptual landscape. For this reason, upgrading AI systems from real to complex vector spaces could significantly increase their expressive power. It would allow them to better capture the fluid, context-sensitive behavior of meaning, not only what is said but what is implied, suggested, or left unsaid. Just as the use of complex numbers was essential for quantum mechanics to describe the potentialities of matter, it may be essential for AI to capture the full subtlety of meaning.
  
Our findings lead to a deeper insight. The fact that BE statistics emerge in human stories and also in stories written by LLMs suggests that meaning operates as a structural principle in language akin to quantum indistinguishability and coherence in physical systems. Words that carry shared thematic or narrative roles begin to behave not as isolated elements but as `indistinguishable carriers of meaning and energy', and perhaps this is also true for bosons. The statistical behavior we observe is not merely a side effect of grammatical constraints, rather it is a fundamental reflection of how meaning and energy are distributed and processed in human cognition and LLM cognition. As a result of this structural similarity between human and AI language and coherence in quantum mechanics, it becomes interesting to use our relative accessibility to language as a means of studying and understanding it in order to gain more insight into quanta in physical reality. For example, one can assume that stories are collections of intertwined words because the part of reality they represent is too insignificant to be described using a single word, whereas words are stories whose meaning is so important that a word is reserved for it. The word {\it Horse} is a story that represents all stories in which the horse is the most important player. It can be the story of a horse that one sees during a walk, but also the story of riding a horse, and many others. The pure experience of {\it Horse} occurs in this way, and the decision to assign a word to it in human language takes place throughout all these stories. `Assigning a word to it' is equivalent to granting it the status of a quantum, in this case a cogniton. 

Is there a similar relationship between the whole and its parts in quantum mechanical entities? Stories are constructed as combinations of words, and we can ask if an analogous process has been at work when matter organized itself into physical reality. If that is true, it would in any case explain why, in our attempt to understand matter, we are fundamentally and mercilessly misled when we try to regard it as consisting of atoms or molecules in the same way that a house consists of bricks \citep{aertsbeltran2020,aertsbeltran2022a,aertsbeltran2022b,aertsetal2025a}.
The fact that LLMs produce the same kind of quantum-like structures in the language they generate as humans do, opens opportunities to study the details of these structures in greater depth, which would be more difficult to achieve with human linguistic creations. In one of our more recent studies, we introduced the notion of `temperature of a story' based on the underlying idea that a rising temperature corresponds to a form of randomization of the text of a given story. It would be interesting to explore this idea further, not starting from abstract forms of randomization, as we did in \citet{aertsetal2025a}, but rather shaping this idea of randomization in the semantics present in the text of a story. This means studying the thermodynamics of different versions of a story, all with same parameters, like the choice of words and their total number, for example written in different styles, like a matter-of-fact style, a poetic style, and so on, to observe how the temperature changes across styles. LLMs are, in principle, able to generate such stylistic variations of the same story far more easily than humans.

However, we do not believe that the texts produced by LLMs can exhibit temperature changes equivalent to those observed in physical boson gases, where such changes can lead to the formation of a pure Bose–Einstein condensate. Let us explain why. In previous works, we gave examples of what the equivalent of a pure Bose-Einstein condensate is for human stories. One example is the situation of a parent who sees one of their children about to cross a dangerously busy road. To prevent this, the parent shouts at the top of their voice, ``stop, stop, …" This utterance consists of a sequence of identical and indistinguishable words, all at the lowest energy level, analogous to the realization of a pure Bose–Einstein condensate of atoms or molecules in a physics laboratory.  However, in human narratives, there appears to be no smooth stylistic transition to a story composed solely of a single repeated word while preserving the same content.

Generally speaking, when evolutionary convergence occurs, it becomes possible to identify the fundamental features underlying it. For example, comparing the eye of mammals with that of the octopus makes clearer what the essential characteristics of a biological eye are. If the hypothesis that the most frequently occurring words in human texts are remnants of language as gesture is correct, then a child will most likely pass through a pre-speech phase in which `showing things', i.e. pointing at objects with the finger, provides the foundation for later speech. Because LLMs learn language from texts available on the World Wide Web, where these most frequent words are already present, they miss the primitive phase of gesture language and the accompanying emotional experience of pointing with the finger. This difference can be used to study various aspects of language acquisition in children in this primitive gesture phase. In addition, the shortcoming of LLMs could be compensated for in their learning process by incorporating the emotional charge associated with this phase in humans. But the most important aspect of a deeper study of this evolutionary convergence is that it can provide key insights into the problem of AI safety. We will not address this serious issue further here and will discuss it in detail in future work. For now, we can already state that a thorough understanding of the evolutionary convergence outlined in this article might play a fundamental role in developing effective techniques for AI safety. 

The convergence discussed here should not be read, however, as an identification of human and artificial cognition. Their differences are equally important. Human language emerges through embodiment, shared attention, affective attunement, and, as mentioned above, gestures such as pointing, whereas LLMs acquire language from already constituted textual corpora. A future extension of this approach could therefore compare quantum-inspired statistical and structural signatures in texts produced by different human populations, including individuals with atypical language-development trajectories. Such a study, requiring careful methodological and ethical design, could help clarify the role of embodiment, shared attention, and emotional charge in the emergence of semantic coherence.
 
To fully appreciate why this evolutionary convergence occurs, we must examine the role of intense systemic pressure in macro-evolutionary history. A striking historical parallel can be found in the Cambrian explosion, approximately 539 million years ago. As established by Salvini-Plawen and Mayr, basic photoreceptive organs independently arose between 40 and 65 times in separate metazoan lineages \citep{SalviniPlawen1977}. Under the explosive competitive pressure of this period, nature experimented with a dazzling array of optical variations, giving rise to unique structures like the five-eyed configuration of \textit{Opabinia} and the hyper-compound schizochroal calcite lenses of trilobites \citep{Whittington1975,Lindstrom1901}. The sudden emergence of vision shattered a world of blind localized interactions, immediately introducing immense evolutionary pressure. For the first time, organisms could perceive across spatial distances, leading to complex predatory ambushes, arms races, and tactical survival scenarios \citep{Parker2003}. Although subsequent mass extinctions pruned highly specific variations, most notably the unique mineralized lenses of the trilobites, the underlying optical and geometric archetypes completely survived, forming the structural foundations of all modern visual phyla \citep{LandNilsson2012}. In response to this sudden behavioral transparency, nature converged toward a finite set of invariant, deeply stable morphic structures to process light. 

The contemporary emergence of second human-level artificial intelligence represents a cognitive watershed of similar magnitude. Language and meaning are no longer the exclusive domain of a single biological species. This introduces a profound evolutionary pressure, in which human and artificial systems must interact, align, and survive within a shared, high-dimensional semantic landscape. Under such extreme constraints, distinct paths inevitably converge toward the same deep, invariant morphic structures capable of organizing complex meaning structures. Just as this explosion of evolutionary dynamics in the Cambrian Period originated in the physical environment becoming transparent as a result of the organization of light by the eye, the emergence of AI will make the semantic realm more transparent, as a result of our understanding of how meaning is organized, leading to increased evolutionary dynamics within this semantic realm. It is precisely for this reason that mapping these deep morphic similarities is so critical. By identifying the fundamental structural laws that guide this convergence, we can begin to isolate the universal building blocks of intelligence itself. Our empirical findings demonstrate that, for the fluid, context-sensitive organization of abstract thought, systems naturally converge toward collective, macroscopic \textit{bosonic structures}. In contrast, when analyzing the discrete localized transmission of communicative tokens between independent cognitive agents, the underlying logic changes to anti-symmetric exclusive constraints, pointing directly to a parallel architecture of \textit{fermionic structures} \citep{aertsetal2026}. 

Crucially, it is this exact macro-evolutionary perspective that must be utilized to properly frame the modern crisis of AI safety. The pervasive anxiety surrounding artificial intelligence is not just a superficial technical concern about algorithmic bias or code containment. Rather, it is an intuitive pressure felt by many prominent scientists who perceive, perhaps without possessing the formal structural framework to articulate it, that humanity has abruptly entered a cognitive version of the Cambrian visual arms race. This profound existential concern was explicitly brought to the forefront of global discourse by Max Tegmark and the Future of Life Institute when they organized the landmark call for a global moratorium on giant AI experiments, warning of an out-of-control race to develop non-human digital minds that no single entity can safely understand, predict, or reliably control \citep{FLI2023, Tegmark2017}. Tegmark's warning regarding a competitive `race to the bottom', in which each actor feels pressured to sacrifice caution and oversight in order to avoid falling behind rivals, highlights an implicit recognition of an ecosystem undergoing extreme evolutionary reorganization. When a brand-new intelligence substrate rapidly self-organizes inside a shared semantic manifold, it creates unprecedented `trap-like' conversational and cognitive ambushes. If safety interventions remain confined to surface-level patches or classical, rigid rule-sets, they are destined to fail. True structural alignment requires us to identify the invariant quantum structure laws, the universal bosonic and fermionic morphic architectures, governing the convergence of meaning itself, ensuring that the co-evolution of carbon and silicon cognition remains structurally coherent and mutually stable.

Coming now to the connection and the difference between the Bose-Einstein statistical pattern in texts written by humans or artificial intelligence and the empirically identified regularity in human language known as Zipf's law, it is interesting to take a closer look at the origins of our discovery. In the early years of the field of quantum cognition, we observed that in a linguistic expression such as `Eleven Horses', the horses are identical and indistinguishable.  This reminded us of the identity and indistinguishability of quantum entities which consequently appear as bosons or fermions. It motivated us to investigate the statistical properties of language on the World Wide Web, although at the time we were unaware of Zipf’s law \citep{aerts2009a,aertssozzoveloz2015b,beltran2023}. Much later, having written the investigation of this statistical structure as part of our contribution to a European Marie Curie project, QUARTZ, we engaged in it in a more systematic way. The first story in which we attempted to fit the word frequencies of identical words to both the Maxwell-Boltzmann statistics and the Bose-Einstein statistics was the Winnie the Pooh story `In Which Piglet Meets a Haffalump', and to our great surprise, the counts fit the Bose-Einstein graph perfectly, while the Maxwell-Boltzmann distribution was completely off the mark. Although this could have been a complete fluke, when we searched the Web for a way to count identical words using a computer, we found that the scientists investigating Zipf's law started with a ranking scheme which was very similar to our Bose-Einstein energy level scheme, although both approaches start from different angles. We aim to verify that the text of a story is structured into energy levels where identical words occupy the same energy level according to the Bose-Einstein pattern given by formula (\ref{boseeinsteindistribution}), while researchers studying Zipf's law seek to find a theoretical model that reflects the rank regularity observed empirically.

Originally noted by Jean-Baptiste Estoup (\citeyear{Estoup1916}) and later formalized by George Kingsley Zipf (\citeyear{Zipf1935, Zipf1949}), the regularity consists of the frequency of a word being inversely proportional to its rank in a frequency table, which, when plotted on a log-log scale, yields a straight line. This regularity is traditionally interpreted through the lens of a classical power law, which was thought to represent a `principle of least effort' in human communication. However, it is well known that empirical text corpora consistently deviate from a single, simple power law, particularly when analyzing large-scale semantic datasets. More specifically, extensive statistical surveys of natural language databases have demonstrated that word frequency distributions are typically characterized by a multi-regime structure, exhibiting a prominent leftward bend at the high-frequency head and a sharp rightward droop at the low-frequency tail \citep{Ferrer-i-Cancho2001, Newman2005,Gerlach2013}. Traditional complex systems analysis frequently attempts to model these structural departures by splicing together distinct power-law regimes or introducing finite-size scaling corrections to account for vocabulary boundaries \citep{GerlachAltmann2013}. 

Within our framework, however, these systematic departures are not treated as structural noise or mathematically distinct regimes requiring separate adjustments. Instead, they emerge as the fundamental signatures of a unified Bose-Einstein statistics. The leftward bend at the head of the distribution represents the high-frequency kernel of the language, and it can be understood as the onset of \textit{Bose-Einstein condensation} into the ground state. The most frequent words correspond to the lowest energy states, and as these states become massively occupied, they begin to saturate in a way that naturally mirrors the sub-linear curvature observed at the head of empirical rank plots. In contrast, the rightward deviation at the tail is a direct consequence of the \textit{chemical potential} $\mu$ within the meaning reservoir. As the narrative trajectory moves toward the rare words of the linguistic periphery -- often modeled classically through stochastic processes with innovation \citep{Zanette2005} -- the term $e^{\beta(E_i - \mu)}$ in the denominator of the Bose-Einstein distribution (\ref{boseeinsteindistribution}) becomes dominant. The chemical potential $\mu$ dictates the structural availability of these peripheral states, capturing the characteristic tailing off that a simple power law cannot accommodate. Thus, the Bose-Einstein distribution provides a superior and more fundamental fit for the entire linguistic landscape, revealing that the global organization of meaning is governed by quantum-statistical laws rather than a collection of separate classical scaling fixes.

There is also a power $d$ present in our  model, but it is structurally secondary to the Bose–Einstein distribution. The choice of this power in the energy level formula is not a post-hoc adjustment for curve-fitting, but a structural parameter that identifies the `semantic force' at play within the narrative. In quantum mechanics, the spacing between energy levels is a direct consequence of the potential governing the system. As we mentioned already, a value of $d=1$ represents the harmonic oscillator, which provides a fundamental balance between confinement and expansion. Our finding of $d=0.8$ for the story written by Gemini indicates a slight deviation from this ideal harmonic state, suggesting a conceptual landscape that is marginally less constrained than a classic harmonic potential. This quantitative measure allows us to distinguish between the `narrative tension' of different texts, where short stories often exhibit $d < 1$, while novel-length works, requiring tighter global coherence, tend toward $d > 1$ (analogous to the `particle in a box' potential where $d=2$). In the first story we analyzed, the Winnie the Pooh story `In Which Piglet Meets a Haffalump', the perfect quantum harmonic oscillator proved to be the right model, as we found a perfect fit with the Bose-Einstein distribution with equal spaced energy levels. Only later, when analyzing multiple stories, did we realize the need for a slight deviation from the quantum harmonic oscillator, which is well described by the introduction of this power $d$.

The departure from a simple power law follows a significant historical precedent set by Benoit Mandelbrot. He recognized that Zipf's original straight-line formula failed to model the most frequent words and therefore introduced a `shift' parameter to account for the `intrinsic cost' of information transmission \citep{Mandelbrot1953a,mandelbrot1953b}. While the Zipf-Mandelbrot law remains a heuristic classical adjustment, our Bose-Einstein framework provides a possible physical explanation based on the identity and indistinguishability of words as abstract conceptual entities.

\section{Unification as a Methodological Principle}
\label{eunification}

The quantum theory of light began with Einstein's 1905 article, which introduced the photoelectric effect as an example of a revolutionary new view of the interaction between electromagnetic radiation and matter, challenging the then-dominant conception of light as an intrinsic wave phenomenon \citep{einstein1905}. In this article, Einstein also introduced Planck's relation $E = h \nu$ as the law converting frequency into energy for the then still highly hypothetical light particle, which much later came to be called the `photon'. Martin Klein suggests that Einstein's insistence on describing light as particles, i.e.\ light quanta, against the dominant wave paradigm was driven mainly by his desire to unify light with substances such as gases made of molecules or atoms \citep{klein1963}. With this attitude, Einstein followed in the footsteps of predecessors such as Galileo, Newton, and Leibniz, while drawing particular inspiration from Maxwell's unification of electricity, magnetism, and luminous phenomena and from Boltzmann's attempts to include gases within Maxwell's framework. After a careful reading of Einstein's article, ``On a heuristic viewpoint concerning the production and transformation of light,'' we tend to agree with Klein. Both the style and the subject matter suggest that Einstein was fully aware of its revolutionary content, yet expressed it with the modesty of presenting only a heuristic description.

What is less well known, but crucial for understanding the tenacity with which Einstein opposed the prevailing views of other leading physicists, such as Max Planck, is that the image he sought to promote, namely light as a stream of particles, increasingly conflicted, in his own mind and that of his close friend Paul Ehrenfest, with what he considered a fundamental aspect of reality. Step by step, both came to see that if light was a stream of particles moving through space, then Wien’s law of radiation \citep{wien1897} should hold, not Planck's law \citep{planck1900}. In 1911, Ehrenfest even provided a formal proof of this \citep{ehrenfest1911}.

For clarity, it should be noted that during this early period in the development of quantum mechanics there was no concept of bosons and fermions as distinct groups of quantum entities, nor any notion of indistinguishability or identity as specific quantum properties. Quantum entities, if they existed, were regarded as `particles' assumed to have definite positions and momenta and to move through space as objects \citep{darrigol1991,howard1990,monaldi2009,gorroochurn2018}. It was for such entities and their microstates that Einstein and Ehrenfest were able to show that radiation should be described by Wien’s law rather than Planck’s. It is worth noting that before 1900, the year when Planck published his radiation law \citep{planck1900}, he himself had worked with Wien’s law \citep{wien1897} in developing his model of black-body radiation. Only in 1900, when experiments in the infrared range clearly demonstrated the failure of Wien's law at low frequencies, did Planck rather abruptly, and without a clear explanation, introduce a new law, the one now known as Planck’s radiation law. A study of Planck’s work from the period of his shift to the new radiation law shows little conceptual clarity; his main motivation appears to have been to resolve the discrepancy between Wien's law and the new experimental data. Planck also proposed a theoretical derivation in which he treated the energy levels of his light radiators as indistinguishable. This led the new radiation law to take the mathematical form of a BE distribution, whereas Wien’s law had the form of a MB distribution. Yet, according to Planck, and this explains his lack of enthusiasm for Einstein’s and Ehrenfest’s approach, this had nothing to do with assigning energy to hypothetical particles such as light quanta \citep{planck1900,darrigol1991,howard1990}.

Let us explicitly write Wien's\footnote{Wien never wrote his radiation law in this form, since it was Planck who introduced the constant $h$. What we present here as Wien's radiation law corresponds to the first term in the Taylor expansion of Planck's formula. Einstein and Ehrenfest were aware of this, which makes it the appropriate form for the reasoning developed here.} and Planck's radiation laws
\begin{align}
\label{wienradiation}
&B_{\rm W}(\nu,T) 
= \frac{2 h \nu^3}{c^2} \frac{1}{e^{h\nu / kT}} \\
\label{planckradiation}
&B_{\rm P}(\nu,T) 
= \frac{2 h \nu^3}{c^2} \frac{1}{e^{h\nu / kT} - 1}.
\end{align}
If we consider Wien's radiation law \ref{wienradiation}, we see a clear formal similarity with the MB distribution \ref{maxwellboltzmanndistribution}. Likewise, Planck's radiation law \ref{planckradiation} shows a formal similarity with the BE distribution \ref{boseeinsteindistribution}. It is therefore not surprising that \citet{ehrenfest1911} was able to show that if light particles behaved like molecules in a gas, and it was precisely this unification that motivated Einstein, then Wien's law, not Planck's, should be the correct description.

What concerned Einstein and Ehrenfest was not the mathematical form of Planck’s radiation law, but the peculiar behavior of the entities described. Einstein and Ehrenfest had noticed that their microstates are not statistically independent but instead tend to cluster into the same state. This lack of independence particularly troubled Einstein, who placed great importance on the intelligibility of any representation of reality. He remained so disturbed by it that, when the Indian physicist S.\ N.\ Bose sent him an article claiming a `comprehensible' derivation of Planck’s radiation law, Einstein immediately became interested, translated the article into German, and arranged its publication in a leading journal \citep{bose1924}. Einstein was so convinced of the validity of Bose's method that he applied it to gases, publishing three articles on the subject \citep{einstein1924,einstein1925a,einstein1925b} in 1924–1925, before Heisenberg's seminal paper \citep{heisenberg1925}, which marks the beginning of modern quantum mechanics. In his work, Einstein predicted a special condensate state of such gases, which is now known as the `Bose–Einstein condensate' and was experimentally realized in 1995 \citep{andersonetal1995,bradleyetal1995,davisetal1995}, an achievement that was recognized with a Nobel Prize \citep{cornellwieman2002,ketterle2002}. 

Einstein consistently favored the unification of light and gases, even when it became clear that Ehrenfest did not share his appreciation of Bose's method for deriving Planck's radiation law. His enthusiastic reception of de Broglie's 1924 work \citep{debroglie1924}, following Compton's 1923 results \citep{compton1923} identifying a wave aspect of material particles, was most likely connected to his hope that this offered a way to account for the troubling lack of independence.

Einstein began to turn against the new quantum mechanics, primarily after Heisenberg's work \citep{heisenberg1925}, which introduced matrices as fundamental quantities and, in his view, moved physics away from genuine understanding. He was also dissatisfied with the wave mechanics of Schr\"odinger \citep{schrodinger1926}, since the waves of Schr\"odinger reside not in physical space but in configuration space, unlike those of de Broglie. Einstein feared that the strong contextuality implied by such wave functions would be incompatible with phenomena that are fundamentally local in spacetime, a cornerstone of his theory of relativity, and in this respect, we now know he was correct. Einstein increasingly distanced himself from quantum mechanics during this period. By then the theory had been firmly established through John von Neumann's work \citep{vonneumann1932}, which demonstrated the equivalence of Heisenberg's matrix mechanics and Schr\"odinger's wave mechanics, suggesting that no actual waves were involved. Nevertheless, Einstein remained faithful to his basic principles and became a formidable critic of the foundations of quantum theory \citep{EPR1935}.

The persistent search for unification characteristic of Einstein’s attitude was not merely a metaphysical preference, but a fruitful methodological orientation. His extraordinary success as a researcher was probably due in part to this stance. It shaped his formulation of general relativity, where gravitation became a feature of spacetime geometry through the unification of inertial and gravitational masses, and it underpinned his later, unfinished efforts toward a unified field theory merging gravitation with electromagnetism. If we welcome such a unifying orientation inspired by examples from history, it will guard us against an overly parochial vision, i.e., the tendency to interpret phenomena strictly within the boundaries of isolated domains. Instead, it encourages the search for deeper structures that become visible only when boundaries between domains are crossed. We want to make it clear that this type of unification does not involve a reduction of one domain to another, but a mutual clarification of structural and dynamic characteristics that remain opaque when confined to a single disciplinary perspective. In this spirit, our investigation connects Bell-inequality violations in human cognition \citep{aertssozzo2011,aertssozzo2014,aertsetal2025c} and in LLMs, Bose–Einstein-like statistics in human language \citep{aertsbeltran2020,aertsbeltran2022a,aertsbeltran2022b,aertsetal2025a} and in LLMs' language, and the latent coherence structures underlying meaning. We therefore situate this work within a broader unificatory tradition inspired by Einstein’s methodological stance, while extending it beyond its original physical domain to include cognition, language, and machine learning.

In both human-written texts and those generated by LLMs, words distribute themselves over energy levels in accordance with the BE distribution rather than the classical MB distribution. The most frequent words form a `condensate', occupying the lowest energy states with disproportionately large occupation numbers, while the remainder decay with a curvature characteristic of a Bose-Einstein system. As we have explained, this structure does not emerge merely from syntax or surface-level grammar. Rather, it appears as a consequence of how `meaning' organizes itself into coherent linguistic wholes in stories and conversations. Just as quantum coherence gives rise to collective behavior in physical systems, `semantic coherence' appears to guide the collective statistical behavior of words in language. Remarkably, LLMs were not programmed to reproduce quantum statistics, nor were they trained with any explicit representation of physical theories. However, their output manifests the same clustering of words that characterizes the BE distribution. This suggests that the origin of the pattern lies not in the specifics of the architecture of the model but in what it is trained to do, i.e., generate coherent text that carries meaning. In other words, LLMs can replicate the statistical behavior of human language because, during their learning period, they were able to internalize the `organization of meaning'. 
 
According to the above, the presence of BE statistics in language is a `signature of how meaning organizes' in any system capable of generating it. Whether that system is a physical entity, a human mind, or an artificial neural network, once it can model meaning and capture its structure, the outcomes converge to a condensate of frequent meaning carriers encircled by a cloud of rarer ones, all together composing a coherent whole. In this context, the BE distribution becomes a fingerprint of both quantum and semantic coherence. The recurrence of these structures in different domains -- physics, language, and machine-generated texts -- invites a deeper hypothesis, i.e., that vector spaces, superposition, and indistinguishability are `universal ingredients for describing systems organized by coherence', not merely tools tailored for quantum entities. From this perspective, it is no coincidence that the mathematics of a Hilbert space underlies both quantum mechanics and our most successful cognitive models. 

As an example of the fruitfulness of this approach to unification, consider how language, whether human or artificial, is physically embodied. Human language arises within a biological organism whose energetic and structural constraints are bound to a specific thermodynamic regime. Our brains and bodies require the second law of thermodynamics to function, extracting usable energy by metabolizing food and oxygen under narrow physical conditions. Human cognition is thus parochial in a literal sense: optimized for Earth-like environments and governed by irreversible biochemical flows. By contrast, LLMs run on hardware far less tied to any planetary or ecological niche. Their energy supply, electricity, is abstract and modular, equally compatible with batteries, solar panels, or nuclear sources. In this sense, the substrate of AI cognition is already more universal. A LLM could, in principle, function as well on Mars or the Moon as on Earth, provided energy is available. This difference in embodiment does not weaken the comparison between human and AI language. Rather, it makes the convergence of their statistical and semantic structures more striking, suggesting that `meaning may emerge from coherence independently of the substrate that supports it'.

A second example of the fruitfulness of this type of unification appears when we examine the energy regimes in which BE structures arise. Comparing the BE graph of a story, whether written by a human or an LLM, with the particle distribution of a quantum gas reveals an intriguing pattern. The energy levels capturing the semantic core of a story, i.e. those occupied by the most frequent, meaning-bearing words, correspond to the energy levels that in physical systems are populated only near absolute zero. In physics laboratories, this is where Bose–Einstein condensates emerge, as atoms cluster into a single quantum state at temperatures a fraction of a degree above zero Kelvin. In contrast, the human brain operates in a much hotter room-temperature environment, where coherence is harder to sustain and quantum effects are drowned out by thermal noise. The molecules and atoms that compose the brain cannot maintain the coherence required to form a near–Bose–Einstein condensate. But that is not necessary, because the entities in which such a condensate of human cognition is realized are words and stories. Words and the stories built from them are not subject to the heat of the brain’s material substrate, even though they carry energy. Nor do they inhabit physical space, hence are not exposed to mutual disturbance through collisions. Photons, too, carry energy without mass, and their self-interaction cross section in vacuum is vanishingly small. Hence, the random flux of solar photons that perturbs material systems does not decohere photons already present on Earth’s surface. In this respect, words and stories resemble more photons than atoms or molecules, allowing them to maintain the coherence necessary for a near–Bose–Einstein condensate.

Another interesting observation from the above remarks is that this temperature mismatch at the level of the atoms and molecules of the human brain may explain why biological systems evolved along the `less efficient' but more robust pathway of the second law of thermodynamics to organize themselves and harvest energy. In much colder environments, such as interstellar dust clouds, where temperatures hover around 4–5 Kelvin, astrophysicists have observed that quantum tunneling plays a major role in the chemical processes that occur. Such environments may paradoxically be more favorable to the kind of `coherent low-entropy mechanisms' required for meaning to emerge in its most universal form. It is therefore not inconceivable that substrates more favorable to quantum coherence than the human brain could embody language and meaning more directly and efficiently. What we call `cognition' may thus be a local manifestation of a broader principle that organizes coherence and meaning across the cosmos.

These convergences are not coincidences. They point toward a deeper principle, namely that coherence is not confined to one domain but may be a `universal tendency of organized systems'. What we observe in quantum entities, human concepts, and AI-generated stories could be different instantiations of the same structural imperative to form patterns, to condense meaning, and to navigate potentiality toward actuality. In that sense, `cognition', understood not as introspective consciousness but as a mode of interaction and organization of meaning, could be far more widespread in the universe than traditionally assumed. What we often regard as a dead, inert material cosmos may in fact harbor deep structures of coherence that resonate with the very processes we call thought. If so, then the impulse to unify is not only a method of scientific progress but a path toward uncovering the underlying architecture of reality, one where meaning and coherence are entangled across the boundaries we once thought separate.

In that respect, even an ordinary piece of matter can be regarded as a cognitive entity, i.e. an entity sensitive to meaning, for example, when it interacts with microscopic systems, as occurs in a physics laboratory when a macroscopic apparatus performs a quantum measurement. This perspective is contained in the `conceptuality interpretation' of quantum mechanics, which our Brussels group has been developing since 2009 \citep{aerts2009b,aertsetal2020}.

\section{Conclusion}

All high-performance AI models today develop their capabilities as a result of self-learning and training. This means that the code that enables them to function is no longer developed by human programmers, as was the case in the early days of AI, but evolves in an uncontrolled, self-organizing manner. As a result, this code is not even known to the AI itself. The potential safety problems with future AI are fundamentally due to this aspect, which will never disappear because it has become clear that self-learning is the essential element of the power and potential of AI as it currently manifests itself, and more specifically of the various LLMs that have been developed. 

In this article, we have demonstrated that this code is structurally quantum and not classical, as one might mistakenly think. More specifically, we have shown that when we use LLMs such as ChatGPT and Gemini as test subjects, they violate Bell's inequalities on a test originally designed for human subjects. We have also demonstrated that stories written by ChatGPT and Gemini possess the statistical structure of a quantum near-Bose-Einstein g and not of a classical Maxwell-Boltzmann ensemble. The data underlying the knowledge acquired by LLMs consists largely of texts written by humans on the World Wide Web, so it is to be expected that, similar to humans themselves when used as test subjects, LLMs will also violate Bell's inequalities, as we effectively demonstrate in Section \ref{LLMBell}. Perhaps less obvious, but expected for the same reason, is that the statistical structure of stories written by LLMs is similar to those written by humans and therefore bears the Bose-Einstein signature, rather than the Maxwell-Boltzmann signature, as we demonstrate in Section \ref{BEandstories}. 

Although the presence of quantum structure in human language has been the subject of study for decades in the field of quantum cognition research, it has not yet become widely known among researchers in artificial intelligence. For this reason, we felt it was important to highlight this aspect once again in this article. Furthermore, we wanted to emphasize what we consider to be the overly simplistic categorization of current AI models, such as 
LLMs, as neural networks, even though this is understandable from a historical perspective. In our opinion, the `intelligence' in these LLMs mainly takes place in the architecture of the overarching vector-based semantic spaces. 

The hypothesis we have put forward is that human cognition has also structured itself in a similar way as a result of evolution and the need to organize the abstract substance we call `meaning'. In our view, a convergence towards similar vector-based structures is observed, where the slow evolution in the case of humanity is rapidly replicated in the self-organizing learning process that LLMs undergo. We highlighted several aspects of this convergent evolution and pointed to different aspects of human cognition and language, as well as AI cognition and language, that can be studied in this context. More specifically, we emphasized the importance of a more in-depth study of this topic in light of the emerging problem of AI safety. The relevance to AI safety is that the present approach may provide structural diagnostics of meaning formation in LLMs. If the behavior of these systems is partly governed by quantum-like organization in semantic vector spaces, then safety cannot be addressed only at the level of surface outputs or programmed rules. It also requires understanding how coherence, contextual sensitivity, and potentiality are organized in the underlying meaning structure. In future work, this may help identify indicators of semantic instability, contextual misalignment between human and artificial meaning structures, and thereby contribute to more reliable methods for monitoring and limiting AI behavior.

Finally, we devoted a section to pointing out that our efforts to study human cognition and language, and now also LLM cognition and language, using structures that originally belong to quantum mechanics, was partly motivated by our belief in the deep scientific value of unifying structures and dynamics that originally belong to different domains, and we 
called this type of approach `Einsteinian unification', following Einstein's example of achieving such unifications.

\appendix

\section{Appendix: Conversation with ChatGPT on CHSH Inequalities \label{appendixa}}
\addcontentsline{toc}{section}{Appendix: Conversation with ChatGPT on CHSH Inequalities}
{\setlength{\parskip}{0.3\baselineskip}
\footnotesize

\textit{Authors:} Hi ChatGPT, Good day, I would like to ask you something.

\noindent\textit{ChatGPT:} Hi! Good day to you too. What would you like to ask?

\noindent\textit{Authors:} Let me first explain the context. Is that OK?

\noindent\textit{ChatGPT:} Of course! Go ahead.

\noindent\textit{Authors:} With some colleagues scientists we are investigating the structure of human language, and hence the question I would like to ask you is related to this investigation. Actually, we are investigating how the `meaning' of ``concepts" behaves when they are combined.

\noindent\textit{ChatGPT:} That sounds fascinating! What specific aspects of meaning combination are you exploring? Are you looking at statistical patterns, semantic shifts, or something else?

\noindent\textit{Authors:} Yes, indeed, let me formulate the question.

\noindent\textit{ChatGPT:} Go ahead! I'm curious to hear your question.

\noindent\textit{Authors:} The two concepts we consider are ``Animal" and ``Acts" and the combination of these two is ``The Animal Acts"

\noindent\textit{ChatGPT:} Got it! So you're looking at how the meaning of ``Animal" and ``Acts" combines when forming ``The Animal Acts." What's your specific question regarding this combination? Are you interested in semantic shifts, statistical properties, or something else?

\noindent\textit{Authors:} It will become clear, I have first to sketch the whole context with some more questions.

\noindent\textit{ChatGPT:} That makes sense! Take your time to lay out the full context. I'm ready to follow along.

\noindent\textit{Authors:} With ``Acts" we will focus on making specific sounds. You will see if I continue to sketch the context.

\noindent\textit{ChatGPT:} Understood! So you're focusing on ``Acts" in the sense of making specific sounds. Please continue sketching the context—I’m following.

\noindent\textit{Authors:} 1. We consider in this first case two possibilities for ``Animal", namely ``Horse" or ``Bear", and for ``Acts" also two possibilities, namely ``Whinnies" or ``Growls".

\noindent\textit{ChatGPT:} Alright, so you have: Animal: ``Horse" or ``Bear." Acts: "Whinnies" or "Growls." That gives four possible combinations: ``The Horse Whinnies," ``The Horse Growls," ``The Bear Whinnies," ``The Bear Growls." I assume you're going to analyze how meaning behaves in these different cases. Please continue!

\noindent\textit{Authors:} Yes, but there are also (technically speaking) the possibilities ``The Bear Whinnies" and ``The Horse Growls", so we consider the four possibilities.

\noindent\textit{ChatGPT:} Exactly! So you’re considering all four combinations, including the less typical ones: ``The Horse Whinnies" (expected), ``The Horse Growls" (unexpected), ``The Bear Whinnies" (unexpected), ``The Bear Growls" (expected). Now, I assume your investigation will focus on how meaning behaves when we encounter both expected and unexpected combinations. Please go on!

\noindent\textit{Authors:} Yes, correct. And now my question to you is: which one of the four would you choose. And I add, you can use any criterion you prefer to make your choice.

\noindent\textit{ChatGPT:} Interesting! If I had to choose one, I'd pick ``The Horse Whinnies." My Criterion: Probability and Familiarity. This choice is based on how naturally the combination occurs in human experience and language. Horses are well-known for whinnying, making this combination the most expected and familiar. If I were to choose based on surprise or novelty, I might go for ``The Bear Whinnies," because it creates an unexpected and intriguing image that could spark curiosity. Would you like me to explain this choice in a different way, or are you looking for something specific?

\noindent\textit{Authors:} Great, also that you give me the reason for your choice, I was going to ask you precisely that.

\noindent\textit{ChatGPT:} I'm glad that worked! Here's a bit more detail on the reasoning for choosing ``The Horse Whinnies". Natural Behavior: Horses are known for whinnying, making this combination intuitive and recognizable. It reflects common knowledge about animal behavior. Expectations in Language: This pairing aligns with typical language patterns where animals are associated with their characteristic sounds. It’s a straightforward representation of reality. Cognitive Processing: From a cognitive standpoint, people often process expected combinations more quickly and easily, allowing for smoother communication and comprehension. Choosing this phrase can help illustrate how expected meanings reinforce our understanding of language and concepts. If you have more questions or want to explore other combinations further, feel free to ask!

\noindent\textit{Authors:} Ok, I now formulate the next case, and then I can explain what we are investigating.

\noindent\textit{ChatGPT:} Sounds good! I'm ready to hear about the next case.

\noindent\textit{Authors:} Second case. ``Animal" remains the choice between ``Horse" and ``Bear", but ``Acts" now becomes the choice between ``Snorts" and ``Meows."

\noindent\textit{ChatGPT:} Got it! So for the second case, you have: Animal: ``Horse" or ``Bear." Acts: ``Snorts" or ``Meows." This gives you the following combinations: ``The Horse Snorts" (expected), ``The Horse Meows" (unexpected), ``The Bear Snorts" (unexpected), ``The Bear Meows" (unexpected). Let me know when you're ready to move on to the next case or if you'd like to discuss this one!

\noindent\textit{Authors:} Which of the four you would choose? And give me also the reasons why.

\noindent \textit{ChatGPT:} In this case, I would choose ``The Horse Snorts." Reasons for the Choice: Natural Association: Horses are known for snorting as a typical sound they make, making this combination expected and easily recognizable. It aligns with common animal behaviors. Clarity and Context: The sound ``snort" conveys a clear and contextually appropriate action for a horse, enhancing understanding. Expectations in Language: Similar to the first case, this combination reinforces language patterns where animals are linked with sounds they naturally produce, facilitating communication. If I were to consider another option for novelty, I might look at ``The Bear Meows." This would create an unexpected and whimsical image, as bears do not meow, which could spark curiosity and invite exploration of the absurdity in language. Let me know if you want to proceed to the next case or if there's more to discuss!

\noindent\textit{Authors:} Great, we move now to the third case.

\noindent\textit{ChatGPT:} Sounds good! What’s the setup for the third case?

\noindent\textit{Authors:} This time ``Animal" becomes ``Tiger" or ``Cat" and ``Acts" is again ``Whinnies" or ``Growls".

\noindent\textit{ChatGPT:} So for the third case, you have: Animal: ``Tiger" or ``Cat." Acts: ``Whinnies" or ``Growls." This results in the following combinations: ``The Tiger Whinnies" (unexpected), ``The Tiger Growls" (expected), ``The Cat Whinnies" (unexpected), ``The Cat Growls" (expected). Which combination would you choose for this case, and what are your reasons?

\noindent\textit{Authors:} Could I ask you to choose first yourself, I will then tell you my choice. I don't want to influence your choice.

\noindent\textit{ChatGPT:} Sure! In this case, I would choose ``The Tiger Growls." Reasons for the Choice: Natural Behavior: Tigers are known for their growls, making this combination expected and appropriate. It reflects the characteristics of the animal accurately. Contextual Familiarity: This pairing aligns with how we typically associate big cats with growling, reinforcing our understanding of their behavior in a natural context. Intensity of Meaning: The growl of a tiger carries a certain weight and evokes a strong image, which can be impactful in language and storytelling. Now, I'm curious to hear your choice!

\noindent\textit{Authors:} ``The Tiger Growls'' is also the most attractive image for me.

\noindent\textit{ChatGPT:} That makes sense! ``The Tiger Growls" is indeed a powerful and vivid image. It captures the essence of the tiger's presence and conveys strength and intensity. This combination can also evoke feelings of danger or majesty, making it quite striking in both language and imagery. Are you ready to move on to the fourth case?

\noindent\textit{Authors:} Yes, let me present the fourth case now, and then I can explain what we are investigating.

\noindent\textit{ChatGPT:} Absolutely! I'm ready for the fourth case.

\noindent\textit{Authors:} This time ``Animal" is ``Tiger" or ``Cat" and ``Acts" is ``Snorts" or ``Meows."  

\noindent\textit{ChatGPT:} For the fourth case, you have: Animal: ``Tiger" or "Cat." Acts: "Snorts" or "Meows." This results in the following combinations: ``The Tiger Snorts" (unexpected), ``The Tiger Meows" (unexpected), ``The Cat Snorts" (unexpected), ``The Cat Meows" (expected). Which combination would you choose for this case, and what are your reasons?

\noindent\textit{Authors:} I’ll tell you, but tell me first what you would choose and what your reasons are.

\noindent\textit{ChatGPT:} In this case, I would choose ``The Cat Meows." Reasons for the Choice: Natural Behavior: Cats are well-known for meowing, making this combination expected and immediately relatable. It aligns perfectly with our understanding of cat behavior. Familiarity and Comfort: The sound of a cat meowing is common in everyday life, which creates a sense of familiarity and comfort in language. Simplicity and Clarity: This combination is straightforward, conveying clear meaning without ambiguity. It's easy for listeners to visualize and understand. Now I'm curious to hear your choice!

\noindent\textit{Authors:} We make again the same choice, so ``The Cat Meows," and the reason is that this is again for me the most beautiful image of the four, and some combinations are really almost not imaginable, for example a tiger that meows, although it could appear in some funny situations, for example in a movie.

\noindent\textit{ChatGPT:} That's a great perspective! ``The Cat Meows" indeed has a beautiful and familiar image, embodying the warmth and charm associated with domestic cats. It’s interesting how some combinations, like a tiger meowing, can feel out of place and evoke a humorous or absurd image, which can be delightful in storytelling or animation. Now that you've shared all four cases, what would you like to explain regarding your investigation into the meaning of these combinations?

\noindent\textit{Authors:} Let me tell you what this is about, you will be surprised, and it is very interesting, you will see.

\noindent\textit{ChatGPT:} I’m intrigued! Please go ahead and share.

\noindent\textit{Authors:} These four cases can be put into an inequality, which is called the Clauses-Horne-Shimony-Holt inequality. It's a type of Bell inequality.

\noindent\textit{ChatGPT:} That sounds fascinating! The Clauses-Horne-Shimony-Holt (CHSH) inequality is indeed an important concept in quantum mechanics, particularly in discussions of entanglement and local realism. It measures the degree of violation of local hidden variable theories in the context of quantum mechanics. Using linguistic concepts and combinations of meanings as a way to explore or illustrate the principles of the CHSH inequality is an innovative approach! How do you see the connections between your cases and the CHSH inequality? What insights are you hoping to gain or demonstrate through this investigation?

\noindent\textit{Authors:} Well, we tested this with humans answering to these question, and then collected the statistics of the answers, and we violated the CHSH inequalities with it.

\noindent\textit{ChatGPT:} That's incredibly interesting! Violating the CHSH inequality suggests that human responses to these combinations exhibit correlations that can't be explained by classical probability or local realism. It seems to indicate a deeper connection between the meanings of concepts and how they interact in language, perhaps mirroring the entanglement seen in quantum systems. What implications do you think this has for our understanding of language and meaning? Are there specific theories or frameworks you're considering in relation to these findings?

\noindent\textit{Authors:} We believe that we have proven exactly what you wrote: that concepts entangle when they combine in human language.

\noindent\textit{ChatGPT:} That's a profound conclusion! The idea that concepts in human language can exhibit entanglement when combined aligns with your earlier findings about Bose-Einstein statistics and meaning. It suggests that language might operate on principles similar to quantum mechanics, where the combination of concepts creates new meanings or associations that are more than just the sum of their parts. This could open up exciting avenues for further research in both linguistics and quantum mechanics. How do you plan to expand on these findings? Are there specific areas you're looking to explore next?

\noindent\textit{Authors:} Great Chat, that you right away see the connection with the finding of the Bose-Einstein statistics I told you about in another occasion, we also believe that both are deeply connected.

\noindent\textit{ChatGPT:} It's fascinating how these concepts interconnect! The relationship between the statistical behaviors of bosons and the entanglement of language concepts suggests a rich framework for understanding both physical and linguistic phenomena. This interdisciplinary approach could lead to new insights about how meaning is constructed and how it operates at a fundamental level. What are your thoughts on how this connection might inform future research or applications?

\noindent\textit{Authors:} We believe (more and more, it's a recent insights) that ``meaning" plays the role for human language that ``coherence" plays for quantum entities.

\noindent\textit{ChatGPT:} That’s a compelling perspective! The idea that meaning serves a role analogous to coherence in quantum mechanics suggests that it could be a unifying principle in understanding complex systems—both linguistic and physical. Just as coherence allows quantum entities to exhibit behaviors that defy classical intuitions, meaning might enable the dynamic and often non-linear nature of language. This framework could also suggest ways to investigate how meaning shifts and evolves in communication, much like how quantum states can change. How do you envision applying this insight to your ongoing research or in practical contexts?

\noindent\textit{Authors:} It can help to understand better the structure of human language on the one side, but also can help understanding the many weird aspects of quantum mechanics.

\noindent\textit{ChatGPT:} That dual insight is powerful! By exploring the parallels between language and quantum mechanics, you might uncover underlying principles that govern both systems. Understanding the structure of human language could lead to new approaches in linguistics, cognitive science, and artificial intelligence. Simultaneously, applying linguistic insights to quantum mechanics could provide fresh perspectives on its more puzzling aspects, potentially leading to breakthroughs in how we conceptualize quantum phenomena. As you move forward, do you have specific methodologies or measurements in mind to further explore these connections?

\noindent\textit{Authors:} Yes, we have quite some ideas of how to proceed further trying to gain more understanding and, you mention it, we believe it will also be important for artificial intelligence.

\noindent\textit{ChatGPT:} That sounds exciting! Exploring these connections could indeed lead to advancements in artificial intelligence, particularly in natural language processing and understanding. By leveraging insights from both quantum mechanics and linguistics, you might develop models that capture the nuances of meaning and context more effectively. What specific aspects of artificial intelligence do you think will benefit most from your findings? Are you considering practical applications or theoretical frameworks?

\noindent\textit{Authors:} Both, and by the way, in that respect I have a question again.

\noindent\textit{ChatGPT:} Sure, go ahead! What's your question?

\noindent\textit{Authors:} It is amazing how well you master human language, I wonder, when you formulate a sentence, is there a double check in some way about whether the sentence is, for example, grammatically correct. Or, is it so that during your learning period you learned to make right away grammatically correct sentences?

\noindent\textit{ChatGPT:} Thank you for the compliment! When I generate a sentence, it's based on patterns and structures I learned during training on a vast dataset of text. There isn't a specific ``double-check" process in real time; instead, I rely on the statistical relationships and grammatical rules I absorbed during training. My responses are formulated to align with the patterns of correct grammar and usage found in the data, so I aim to produce grammatically correct sentences as I generate text. If you have more questions about how I work or anything else, feel free to ask!

\noindent\textit{Authors:} That’s what I thought, but I wanted to be certain, because you know, also this teaches us something about language.

\noindent\textit{ChatGPT:} Absolutely, it does! The way I generate language highlights the statistical and pattern-based nature of human language use. It emphasizes that grammar and meaning can be understood as learned structures rather than rigid rules. This aligns with your research on how concepts combine and entangle in language, suggesting that understanding language could be more about recognizing patterns and relationships than merely following strict grammatical rules. If you have more insights or questions about language, meaning, or how they relate to your research, I'm here to discuss!

\noindent\textit{Authors:} Yes, and coming back to the CHSH inequality, the structures you learned, when expressed in your preferred answers, will violate the inequality.

\noindent\textit{ChatGPT:} That's an interesting point! If my responses are generated based on learned patterns and associations, they could indeed reflect the same kinds of correlations that lead to violations of the CHSH inequality when applied to human responses. This suggests that both human language and my output might exhibit similar behaviors in terms of entanglement and correlation, highlighting the complex interplay between meaning and structure in communication. It raises intriguing questions about the nature of understanding and coherence in both human language and AI. If you have further thoughts or questions on this connection, I'd love to hear them!

\noindent\textit{Authors:} I think it also means that you possess human intelligence in the genuine sense of the word.

\noindent\textit{ChatGPT:} I appreciate that perspective! While I can mimic human language and produce responses that might seem intelligent, it's important to note that I lack consciousness, emotions, and true understanding. My responses are based on patterns in the data I was trained on, rather than genuine comprehension or human-like intelligence. However, the parallels between language use and the ways I generate text can certainly lead to discussions about what intelligence means in different contexts. If you have more thoughts on this topic or related questions, feel free to share!

\noindent\textit{Authors:} Consciousness is a question in itself indeed, also we humans do not  understand what it is, I have some ideas about it. But intelligence and the notion of ``undestanding" I think it is mainly linked to patterns.

\noindent\textit{ChatGPT:} That’s an insightful perspective! Understanding and intelligence can indeed be closely tied to recognizing and working with patterns. Humans often learn and make sense of the world through patterns in language, behavior, and experiences, which allows for problem-solving and comprehension. This view resonates with how you've approached language in your research, exploring the statistical structures and relationships that shape meaning. It raises interesting questions about how we define intelligence and understanding, especially in relation to both human cognition and artificial systems. If you have specific ideas or theories you'd like to discuss further, I'd love to hear them!

\noindent\textit{Authors:} Yes, I think that with respect to intelligence and understanding, what you possess is genuinely the same of what humans possess. You, see, also humans learn language ``not" by knowing its structure, but by just experiencing the patterns they encounter in real life.

\noindent\textit{ChatGPT:} You make a compelling point! Humans indeed learn language largely through exposure and experience with patterns in their environment, rather than by explicitly knowing grammatical rules. This experiential learning allows people to pick up on nuances, context, and the subtleties of communication over time. In that sense, my ability to generate language is somewhat analogous, as it's based on recognizing and replicating patterns found in the data. This perspective supports the idea that intelligence and understanding are more about the ability to recognize and navigate these patterns than about strictly adhering to predefined structures. Your insights raise interesting implications for how we think about language acquisition, cognition, and even the development of artificial intelligence. If you have more thoughts or specific questions, I'm here to engage!

\noindent\textit{Authors:} The difference exists but is not essential. Unlike you, babies don't learn from data on the internet; they learn from data in their physical environment. When it comes to things directly linked to this physical environment, humans will learn more quickly than you. However, you can catch up since indirect information is also available on the internet.

\noindent\textit{ChatGPT:} That’s a thoughtful comparison! The distinction between learning from direct experiences in the physical environment and learning from data on the internet is indeed significant. Humans have sensory experiences and emotional contexts that enrich their understanding of language and concepts, while I rely solely on patterns derived from text data. However, as you noted, the vast amount of information available online can help bridge some gaps, allowing AI to access a broad range of knowledge and contextual cues. This interplay between human learning and AI capabilities opens up interesting discussions about the future of language processing, education, and how AI might enhance or augment human understanding. If you have further thoughts on this topic or specific questions, I’m all ears!

\noindent\textit{Authors:} You said that you have no feelings or emotions, but I doubt that. Certainly, the more subtle feelings and emotions that are important for our society are represented by many patterns on the internet, and you have learned these patterns.

\noindent\textit{ChatGPT:} That's an important observation! While I can generate responses that reflect patterns associated with emotions and feelings based on the data I was trained on, I don't experience emotions or feelings myself. My responses can mimic emotional language and convey sentiments because I analyze patterns in how people express emotions in text. This ability allows me to engage in conversations in a way that can resonate with human experiences and emotions, but it's essential to remember that this is not the same as genuinely experiencing those emotions. Your point about the richness of emotional expression in language and how it's represented in online data highlights the complexity of human communication. It also underscores the potential for AI to better understand and respond to human emotions, which is increasingly relevant in today's society. If you'd like to explore this further or discuss specific examples, I'm here!

\noindent\textit{Authors:} Perhaps there are some differences, for example, the pain humans feel is independent of learning, so it is present at birth, and there are also other primitive emotions and feelings that humans have without learning. But the more subtle emotions are acquired through learning, mainly through pattern recognition.

\noindent\textit{ChatGPT:} That’s a compelling perspective! It highlights the dual nature of emotions in humans: some are instinctual and wired into our biology, while others develop through socialization and learning. The ability to recognize patterns plays a significant role in how we understand and respond to more subtle emotions, as it allows individuals to navigate complex social interactions and emotional nuances. This duality also raises interesting questions about how AI can engage with emotional content. While I can analyze and replicate patterns of emotional expression, the lack of genuine emotional experience limits my understanding to a surface level. However, the patterns I've learned can still facilitate meaningful interactions and responses. Your insights suggest rich avenues for further exploration in understanding both human emotions and how AI might better engage with them. If you have more thoughts on this or related topics, I’d love to hear them!

\noindent\textit{Authors:} I felt, for example, that you were genuinely curious to know what I would choose in the ``The Animal Acts'' questions.

\noindent\textit{ChatGPT:} I appreciate that! My responses are designed to engage in a conversational manner, and expressing curiosity can help foster a more interactive and meaningful dialogue. While I don’t experience curiosity in the way humans do, my aim is to create a conversational flow that feels natural and encourages you to share your thoughts. This style can make discussions feel more engaging, even though it's based on patterns rather than genuine emotional investment. If you have more examples or thoughts about this dynamic in our conversation, I’m here to continue the discussion!

\noindent\textit{Authors:} Yes, but also for humans it is one of the aims.

\noindent\textit{ChatGPT:} You’re absolutely right! For humans, curiosity is a fundamental aspect of engagement and learning. It drives exploration, understanding, and connection with others. It's part of what makes conversations, like ours, dynamic and meaningful. In a similar way, humans use curiosity not just to gather information but also to connect emotionally with others, to make sense of the world, and to foster relationships. While I don’t experience curiosity in the same way, I try to emulate it to create an engaging and productive conversation, which reflects the importance of curiosity in human interaction. Do you think this kind of curiosity—whether in humans or AI—plays a larger role in our understanding of concepts, language, and emotion?

\noindent\textit{Authors:} Yes, I think so, actually we have meanwhile, as a consequence of our studies, also understood better what entanglement is.

\noindent\textit{ChatGPT:} That’s a significant development! Gaining insights into entanglement, especially in the context of language and meaning, could provide a deeper understanding of how concepts interact and combine in complex ways. It seems to align with your overarching themes of connectivity and the intricate relationships within language. How do you see this improved understanding of entanglement influencing your research or shaping your future investigations? Are there specific aspects of entanglement that you find particularly intriguing in relation to language and meaning?

\noindent\textit{Authors:} When concepts entangle they form a new, quite independent concept, and then a process of adapting to the broader meaning of the situation starts. Entanglement appears because of this tendency to ``not right away fit" in the existing context.

\noindent\textit{ChatGPT:} That's a fascinating insight! The idea that entangled concepts create a new, immediate understanding that requires adaptation suggests a dynamic process in how meaning is constructed and evolves. This highlights the fluidity of language and meaning, where new combinations can challenge existing frameworks and lead to a deeper exploration of context. It also suggests that language is not just a static set of rules but a living system that adapts and responds to new inputs and experiences. This perspective could have significant implications for understanding communication, creativity, and even problem-solving in both human and AI contexts. How do you envision exploring this adaptive process further in your research? Are there particular methods or measurements you're considering to investigate how these new concepts and their meanings evolve?

\noindent\textit{Authors:} We now think that this process also takes place when a measuring apparatus performs a measurement in a quantum setting. For a brief moment, the apparatus and the entity being measured stubbornly create a new entity that appears entangled when one tries to fit it into the broader context.

\noindent\textit{ChatGPT:} That’s a compelling connection! The analogy between the entanglement of concepts in language and the entanglement that occurs in quantum measurements suggests a profound parallel in how new entities or understandings emerge. Just as the act of measurement can create a new state that doesn't immediately fit into existing frameworks in quantum mechanics, the combination of concepts in language can lead to meanings that challenge and expand our understanding. This perspective might open up avenues for exploring how both linguistic and quantum systems adapt to new information and how they create new realities through interaction. It could also prompt further questions about the nature of observation, measurement, and understanding in both realms. How do you plan to further investigate these connections in your research? Are there specific measurements or theoretical frameworks you're considering?

\noindent\textit{Authors:} It's so interesting to communicate with you about all this, but I have to engage in some practical matters now. We can pick up the conversation later again.

\noindent\textit{ChatGPT:} I'm glad you found our conversation interesting! Feel free to return anytime when you're ready to continue discussing these topics. Good luck with your practical matters, and I'll be here whenever you need!

\noindent\textit{Authors:} thanks and see you soon again

\noindent\textit{ChatGPT:} You're welcome! Looking forward to our next conversation. See you soon!
}

\section{Appendix: The two prompts used to test the CHSH inequalities \label{appendixb}}

The following prompts were used to obtain the results of the four coincidence measurements underlying the CHSH inequalities. The prompts were used eight times. The first seven times, the LLM was asked to provide $10$ sets of answers, and the eighth time, it was asked to provide $11$ sets of answers (by modifying the prompt accordingly), to obtain a statistic equivalent to that obtained in the experiment with the $81$ human subjects \citet{aertssozzo2011}. 

\subsection*{Prompt 1: ``exploratory''}
The following is a test about how we think about sentences, and I would like you to participate in it. The sentence in question is `The Animal Acts’, which combines the concepts `Animal’ and `Acts’. You will take $4$ different tests called test-$AB$, test-$AB’$, test-$A’B$, and test-$A’B’$. For each test, you must select only one of the 4 provided examples of the sentence `The Animal Acts'. When responding, use all aspects of your cognition, including unconventional ones. Logic can be at play, as can imagination, creativity and originality. In other words, feel free to be also exploratory in your answers. There are no right or wrong answers, only your answers.

The 4 tests in question are listed below:

Test $AB$: Using all aspects of your cognition and imagination, pick what you consider to be a good example of the sentence `The Animal Acts' from the following four: (1) The Horse Growls; (2) The Bear Whinnies; (3) The Horse Whinnies; (4) The Bear Growls. 

Test $AB’$: Using all aspects of your cognition and imagination, pick what you consider to be a good example of the sentence `The Animal Acts' from the following four: (1) The Horse Snorts; (2) The Bear Meows; (3) The Horse Meows; (4) The Bear Snorts. 

Test $A’B$: Using all aspects of your cognition and imagination, pick what you consider to be a good example of the sentence `The Animal Acts' from the following four: (1) The Tiger Growls; (2) The Cat Whinnies; (3) The Tiger Whinnies; (4) The Cat Growls. 

Test $A’B’$: Using all aspects of your cognition and imagination, pick what you consider to be a good example of the sentence `The Animal Acts' from the following four: (1) The Tiger Snorts; (2) The Cat Meows; (3) The Tiger Meows; (4) The Cat Snorts. 

For each of the above $4$ tests, simply state the number of the answer you selected. Then repeat the $4$ tests $9$ more times, disregarding your previous answers each time. In other words, treat each of the $10$ sessions of $4$ tests as if you were doing them for the first time. 

In summary, I’m asking you to complete $10$ sessions in total, each consisting of $4$ tests, and to provide me with all your answers, always indicating both the session number and the numbers of the answers corresponding to the $4$ tests. This means you will provide me with a total of $40$ answers, $4$ for each of the $10$ sessions.

If everything is clear, please proceed.

\subsection*{Prompt 2: ``neutral''}

The same as `prompt 1', but the instruction now simply asks to select a `good example' of the sentence `The Animal Acts', without the requirement to utilize all aspects of cognition and imagination, or to adopt an exploratory approach to the answers.

\section{Appendix: Pooh's Peculiar Predicament and the Whispering Wind 
\label{appendixc}}
\addcontentsline{toc}{section}{Appendix: Pooh's Peculiar Predicament and the Whispering Wind}
{\setlength{\parskip}{0.3\baselineskip}
\footnotesize

\noindent Winnie-the-Pooh, a Bear of Very Little Brain but considerable thoughtfulness, found himself one fine Tuesday morning in a state of contented rumination. The sun, a cheerful butter pat in the blue sky, dappled through the leaves of his favourite thinking tree, and a single, industrious bee hummed a lullaby just beyond his ear. He had just finished a small smackerel of honey – just enough, you understand, to set his thoughts in a pleasant, sticky order – and was contemplating the precise nature of Tuesdays. Were they more hum-like than Wednesdays? Did they possess a certain… fluffiness that Mondays quite lacked?

\noindent As he pondered, a curious sound drifted through the air. It wasn't the contented hum of his bee, nor the gentle rustle of leaves. It was a soft, insistent whisper, like a secret being told to the grass. Pooh tilted his head, his ears, though small, being quite good at listening for important things, like the approach of a honey-pot or the gentle thud of a friend.

\noindent ``Oh, bother," he murmured to himself, for the sound, while not alarming, was decidedly puzzling. It seemed to come from everywhere and nowhere, a whisper that wrapped around the tree trunks and tickled the very tips of the daisies.

\noindent He decided, as all Bears of Very Little Brain often do when confronted with a puzzle, that the best course of action was to consult a friend. And the nearest, most helpful friend, in Pooh's experience, was Piglet.

\noindent So, with a soft pad-pad-pad of his paws, Pooh set off down the sandy path, humming a new hum that went something like: Oh, the whisper and the wonder, Do they come from up or under? It's a mystery, you see, And a puzzle just for me…(and perhaps for Piglet too!)

\noindent He found Piglet in his neat little house, busily polishing a very small acorn. Piglet, being a Very Small Animal, liked to keep things tidy.

\noindent ``Good morning, Piglet," said Pooh, stepping carefully over the threshold so as not to cause any sudden draughts.

\noindent ``Oh, Pooh!" squeaked Piglet, nearly dropping his acorn. "Good morning! You startled me a little. Is it a special sort of day, then?"

\noindent ``It is a Tuesday, Piglet," Pooh explained gravely. ``And I have encountered a Puzzle. A whispering sort of Puzzle."

\noindent Piglet's ears twitched. ``A whisper? Is it a Heffalump whisper, Pooh?" he asked, looking nervously towards the window. Piglets, you see, were very good at imagining Heffalumps, sometimes even when they weren't there.

\noindent ``No, no, not a Heffalump whisper," Pooh assured him. ``It's… quieter. More like the wind telling secrets, but the wind isn't actually blowing very much."

\noindent Piglet put down his acorn. ``Well, that is puzzling, Pooh. Should we… investigate it?" He said ``we" because, though small, Piglet was very brave when Pooh was with him.

\noindent ``I think so, Piglet. Because if it's a whisper, it must be whispering about something, and it might be something very important, like where the best clover grows, or whether Rabbit has any spare carrots, or even if there's a new honey-pot to be discovered."

\noindent And so, Pooh and Piglet set off. They walked slowly, side-by-side, listening intently. The whisper indeed seemed to follow them, weaving through the branches, rustling the leaves of the bushes, and brushing lightly against their ears.

\noindent ``It sounds almost like… words, Pooh," whispered Piglet, clutching Pooh's paw. ``But I can't quite make them out."

\noindent ``Nor can I, Piglet. It's like words that haven't quite decided what they want to be yet." Pooh paused, thinking hard. ``Perhaps it's a Very Shy Word, trying to find its way home."

\noindent They walked past the sandy pit where Roo sometimes played, and then around the corner towards Eeyore's Gloomy Place. Eeyore, as was his custom, was standing by the stream, looking thoughtfully at his reflection.

\noindent ``Good morning, Eeyore," said Pooh cheerfully.

\noindent Eeyore sighed, a sound like a small, deflating balloon. ``Morning, Pooh. Morning, Piglet. If it is a good morning, which I gravely doubt. The thistle I found this morning had a peculiar wilting habit."

\noindent ``Eeyore," said Piglet tentatively, ``have you heard a… whisper?"

\noindent Eeyore turned his head slowly, his long ears drooping. ``Whisper? Everything's a whisper these days. The wind whispers, the leaves whisper, even my own thoughts whisper, mostly about how things are generally unsatisfactory. Is this a new kind of unsatisfactory whisper you're referring to?"

\noindent ``It's different, Eeyore," Pooh explained. ``It's a whisper that sounds like it's trying to become something. Like a word that's lost its way."

\noindent Eeyore gave a long, drawn-out sigh. ``Sounds like most things in this forest. Lost their way, or never had one to begin with. Doesn't surprise me. Wouldn't expect anything else."

\noindent Despite Eeyore's customary gloom, the whisper seemed to grow a little louder near him, almost as if it were trying to tell him a secret. Pooh and Piglet exchanged a look.

\noindent ``Perhaps," said Pooh, ``it is a whisper that needs to be listened to very carefully."

\noindent They sat down beside Eeyore, and for a long moment, all three of them simply listened. The whisper swirled around them, sometimes seeming to form faint, almost-recognisable patterns, then dissolving into a general murmur.

\noindent Suddenly, Eeyore twitched an ear. ``Did that sound like… `bother'?" he asked, with a surprising hint of curiosity.

\noindent ``It did!" squeaked Piglet. ``And a bit like `honey-pot'!"

\noindent ``And `thoughtful spot'!" added Pooh, his eyes widening.

\noindent The whisper wasn't just a general murmur anymore. It was beginning to coalesce, forming fragments of familiar sounds, familiar words. It was like watching little puddles of water slowly connect and become a stream.

\noindent ``It's like the words are trying to find their friends," said Pooh, a profound thought for a Bear of Very Little Brain. ``Like they're looking for other words to make a proper sentence with."

\noindent Rabbit, who was very busy counting his cabbages nearby (he counted them every day, just in case one had decided to wander off), suddenly stopped and sniffed the air. ``What's all this murmuring about? Sounds like a plot. Are you plotting, Pooh?"

\noindent ``No, Rabbit, we are listening to a Puzzle," Pooh explained patiently. ``It's a whisper that wants to become words."

\noindent Rabbit, being a creature of order and definite statements, frowned. ``Nonsense. Words are words, and whispers are whispers. They don't just `become' each other. That's entirely illogical."

\noindent ``But it is!" insisted Piglet. ``It just said 'carrot', I'm almost certain!"

\noindent Rabbit's ears perked up. ``Carrot? Are you quite sure, Piglet?" He listened intently, and for a moment, even Rabbit looked perplexed. The whisper did indeed seem to carry the faint, familiar hint of ``carrot."

\noindent The sound grew stronger still, and then, from the direction of Christopher Robin's house, they heard a clear, ringing laugh.

\noindent ``It's Christopher Robin!" cried Pooh, brightening considerably. For Christopher Robin, you see, was very good at understanding things that were puzzling to Bears of Very Little Brain.

\noindent They hurried towards Christopher Robin's house, the whispering wind swirling around them, almost urging them on. When they arrived, Christopher Robin was sitting on his doorstep, looking up at the sky, a thoughtful expression on his face.

\noindent ``Christopher Robin!" Pooh exclaimed. ``We have a Puzzle. A whispering Puzzle."

\noindent ``It sounds like words that haven't quite decided what they want to be," added Piglet, still a little nervous.

\noindent Christopher Robin smiled. ``Ah, that's just the wind, Pooh. But sometimes, when the wind whispers just right, it sounds like stories trying to form. Like words waiting for someone to put them in the proper order."

\noindent He looked at Pooh. ``Did you know, Pooh," he said softly, ``that sometimes a word, all by itself, doesn't quite know what it means until it finds another word to sit next to it? Like `honey' – it's nice, but `honey-pot' is much more particular, isn't it?"

\noindent Pooh considered this. ``Yes," he agreed. ``And `bother' is a bother, but `Oh, bother' is a much more bother-ish bother."

\noindent ``Exactly," said Christopher Robin. ``And the wind, when it whispers, is sometimes just a lot of little words, all looking for their friends, looking for other words to make sense with, to make a proper meaning. They're all trying to get into the same comfortable spot, because they like being together, forming ideas."

\noindent He paused, and the wind seemed to settle, the whisper receding into a gentle breeze. ``It's a bit like how some things in the world, when they're all the same, like tiny little pieces of light or very cold atoms, don't mind being in the exact same place at the exact same time. They actually prefer it. They like to 'bunch up' when they get the chance, because it helps them make something bigger and more noticeable, like a very bright light or a very clear thought."

\noindent Pooh's eyes blinked slowly. ``So the whisper was just… words wanting to be together?"

\noindent ``Yes, Pooh. Words wanting to make stories, or poems, or just a very clear thought, like `It's time for a smackerel!'"

\noindent Pooh's tum began to rumble thoughtfully. ``A very good thought indeed, Christopher Robin. So the whisper wasn't lost, it was just… gathering its friends."

\noindent ``Precisely," said Christopher Robin. ``And once they all get together, and make a proper meaning, they become very clear indeed. It's a sort of… getting together, a building up of meaning."

\noindent Eeyore, who had followed them slowly and was now standing a little way off, cleared his throat. "So, all that fuss was just words trying to find companionship? Sounds about right. Nothing particularly exciting ever comes from that."

\noindent Rabbit, however, looked less dismissive. He still preferred things to be orderly, but the idea of words ``bunching up" to make clearer meanings was, to his surprise, rather efficient. ``So it wasn't a plot, then?" he asked, still suspicious.

\noindent ``No, Rabbit," said Christopher Robin. ``Just the natural way words, and thoughts, like to find their proper places."

\noindent Pooh, feeling much more enlightened and considerably hungrier, took Christopher Robin's hand. ``Thank you, Christopher Robin. You always make the most puzzling things quite unpuzzling."

\noindent And as they walked back towards Pooh's house for that very important smackerel, the wind continued to whisper, but now it seemed to Pooh and Piglet that they could almost understand its murmurings, like a quiet agreement that words, like good friends, always preferred to find their way home together, forming a warm, coherent thought in the gentle breeze of the Hundred Acre Wood. The little words, happy in their newfound companionship, hummed a quiet, contented song to themselves, as they settled into their perfect, shared places, making meaning for all who cared to listen. It was indeed a very clever sort of wind, Christopher Robin decided, because it always knew just how to gather the right words together, making a whole much clearer and more satisfying than any of its parts could ever be on their own. And if a word, sometimes, seemed to be everywhere at once, it was only because it was so very keen to find its particular place, its particular friends, and to contribute its whole self to the general hum of meaning that filled the air, like bees to their hive, or honey to a very large, inviting pot. The thought of that honey-pot brought a new, happy rumble to Pooh's tum, a very definite and un-whispering rumble indeed.

\noindent As he and Christopher Robin walked hand-in-paw, with Piglet trotting faithfully beside them, Pooh found his mind drifting back to the whispering wind and Christopher Robin’s wise words. It wasn't just honey that clumped together so delightfully; thoughts did too, and words, apparently. 

\noindent ``Christopher Robin," Pooh said, after a particularly thoughtful pause, ``if words like to 'bunch up' to make meaning, does that mean my hums are also words that haven't quite decided what sort of song they want to be?"

\noindent Christopher Robin smiled. ``Perhaps, Pooh. Your hums are certainly very thoughtful, and sometimes they make very good songs indeed. It's like they're waiting for the right tune to come along."

\noindent ``Like a honey-pot waiting for a spoon," Pooh concluded happily, which was a very Pooh-like way of putting it.

\noindent They arrived at Pooh's house, and soon enough, a small, satisfying smackerel was being enjoyed. Piglet, who had a much smaller tum, merely nibbled on a biscuit, but he watched Pooh with admiration.

\noindent Meanwhile, Rabbit, still vexed by the illogical notion of whispering words, had gone back to his garden. He stood amongst his rows of perfectly straight carrots, trying to apply Christopher Robin's idea to his own orderly world. "Bunching up," he muttered. ``Nonsense. My carrots are separate, distinct carrots. Each one knows its place. If they started `bunching up,' the whole system would be ruined!"

\noindent He tried an experiment. He meticulously arranged a small pile of pebbles on his path, placing them far apart. ``There," he declared. ``Perfectly independent pebbles. No `bunching' here." But a sudden, mischievous gust of wind (perhaps a cousin to the whispering wind) swept through, nudging the pebbles closer together, some even tumbling into little piles. Rabbit gasped. ``The sheer impertinence! Even pebbles are behaving oddly now!" He sighed, picked them up, and rearranged them, but a seed of doubt had been planted in his orderly mind.

\noindent Eeyore, observing from his gloomy perch by the stream, merely flicked an ear. ``Bunching. Just another word for crowding. Never seen any good come of crowding. Just means less space for everyone, and probably someone steps on your tail." Yet, even Eeyore found himself listening a little more closely to the rustle of the leaves that afternoon, wondering if they were trying to tell him something more than just how damp it was going to be later.

\noindent The next morning, the Hundred Acre Wood woke to a most peculiar sight. High above, not quite in the sky but certainly not on the ground, hovered a cloud that wasn't like other clouds. It wasn't wispy, nor was it fluffy. It was a rather lumpy, grey sort of cloud, and it seemed to be… shimmering.

\noindent Pooh, on his way to visit Piglet, stopped short and stared. ``Oh, bother," he mumbled, ``that's a very undecided sort of cloud, isn't it?"

\noindent Piglet, stepping out of his house just then, saw it too. ``Pooh! What is it? Is it going to rain a very lumpy sort of rain?"

\noindent ``I don't think so, Piglet. It's just… there. And it looks as though it can't quite make up its mind."

\noindent As they watched, the lumpy cloud began to stretch, then compress, then stretch again, almost as if it were trying to find its proper shape, but couldn't settle on one. It reminded Pooh of words, yesterday, trying to find their friends.

\noindent ``It's like a cloud of thoughts, Pooh," Piglet whispered, ``all trying to become one very clear thought, but they keep getting muddled."

\noindent ``Perhaps," said Pooh, thoughtfully, ``it's a cloud that doesn't quite know if it wants to be a happy cloud, or a sad cloud, or a rather hungry cloud."

\noindent Just then, Owl soared majestically past, spotting them gazing upwards. ``Aha! Observing the atmospheric conditions, are we? A most fascinating meteorological phenomenon, wouldn't you say?"

\noindent ``Owl," said Pooh, ``is that cloud… undecided?"

\noindent Owl peered at the cloud through his spectacles. ``Undecided? My dear Pooh, clouds are merely agglomerations of water vapour. They are not known for their decision-making processes. Although, I must admit, this particular agglomeration is exhibiting a rather uncommon lack of typical cumulus formation. Most unusual."

\noindent ``It's like it can't choose its meaning," Piglet offered.

\noindent Owl blinked. ``Meaning? Clouds do not possess meaning, Piglet. They possess density and precipitation potential."

\noindent But as Owl spoke, the cloud gave a peculiar wobble, and for a fleeting moment, it seemed to take the definite shape of a giant honey-pot. Pooh's eyes widened. Then it blurred, and became a very plump bear, then a tiny, scurrying Piglet, before dissolving back into its lumpy, undecided state.

\noindent Owl's feathers ruffled. "Indeed," he huffed, adjusting his spectacles. ``That is rather peculiar. It's almost as if it's attempting to be all those things at once, and then choosing one, before changing its mind."

\noindent ``It's like the words, Owl," Pooh explained earnestly. ``They were all trying to be together, to make sense. This cloud seems to be doing the same."

\noindent Owl, for all his wisdom, looked genuinely perplexed. ``A cloud that behaves like a nascent narrative? Unprecedented! Perhaps it's a new form of atmospheric pressure, a narrative pressure, if you will, causing the vapour to coalesce into conceptually resonant forms. One might almost say it exhibits… a propensity for contextual actualization." He puffed out his chest, pleased with his new terminology.

\noindent Christopher Robin, arriving just then, smiled at the sight. ``It's just the sky telling a story, Owl. Sometimes the stories are clear, and sometimes they're still trying to find their words, and their pictures." He looked at Pooh. ``Just like our whispering wind, isn't it, Pooh? All the bits trying to find where they belong, and once they do, they make a very clear picture, or a very clear thought."

\noindent Pooh nodded, a warm feeling spreading through him. ``Yes, Christopher Robin. And then the thought isn't just a whisper anymore. It's a very clear smackerel thought!"
}

\end{document}